\documentclass{article}

    \PassOptionsToPackage{numbers, compress}{natbib}

\usepackage{xcolor}

    \usepackage[final]{neurips_2020}

\usepackage[utf8]{inputenc} 
\usepackage[T1]{fontenc}    
\usepackage{hyperref}       
\usepackage{url}            
\usepackage{booktabs}       
\usepackage{amsfonts}       
\usepackage{nicefrac}       
\usepackage{microtype}      
\usepackage{amsmath,amssymb} 
\usepackage{color}
\usepackage{multirow}
\usepackage{colortbl}
\usepackage{graphicx}
\usepackage{epsfig}
\usepackage{grffile}
\usepackage{wrapfig}
\usepackage{subcaption}
\usepackage{changes}
\usepackage{enumitem}

\newcommand\norm[1]{\left\lVert#1\right\rVert}
\newcommand{\normone}[1]{\Vert #1 \Vert_1}

\newtheorem{theorem}{Theorem}

\title{Distribution Matching for Crowd Counting}

\author{%
  Boyu Wang$^{*}$ \quad Huidong Liu$^{*}$ \quad Dimitris Samaras \quad Minh Hoai \\
  Department of Computer Science, Stony Brook University, Stony Brook, NY 11790 \\
  \texttt{\{boywang, huidliu, samaras, minhhoai\}@cs.stonybrook.edu}\\
   $^{*}$indicates equal contribution
}

\begin{document}
\def\mA{\mathcal{A}}
\def\mB{\mathcal{B}}
\def\mC{\mathcal{C}}
\def\mD{\mathcal{D}}
\def\mE{\mathcal{E}}
\def\mF{\mathcal{F}}
\def\mG{\mathcal{G}}
\def\mH{\mathcal{H}}
\def\mI{\mathcal{I}}
\def\mJ{\mathcal{J}}
\def\mK{\mathcal{K}}
\def\mL{\mathcal{L}}
\def\mM{\mathcal{M}}
\def\mN{\mathcal{N}}
\def\mO{\mathcal{O}}
\def\mP{\mathcal{P}}
\def\mQ{\mathcal{Q}}
\def\mR{\mathcal{R}}
\def\mS{\mathcal{S}}
\def\mT{\mathcal{T}}
\def\mU{\mathcal{U}}
\def\mV{\mathcal{V}}
\def\mW{\mathcal{W}}
\def\mX{\mathcal{X}}
\def\mY{\mathcal{Y}}
\def\mZ{\mathcal{Z}}

\def\1n{\mathbf{1}_n}
\def\0{\mathbf{0}}
\def\1{\mathbf{1}}

\def\A{{\bf A}}
\def\B{{\bf B}}
\def\C{{\bf C}}
\def\D{{\bf D}}
\def\E{{\bf E}}
\def\F{{\bf F}}
\def\G{{\bf G}}
\def\H{{\bf H}}
\def\I{{\bf I}}
\def\J{{\bf J}}
\def\K{{\bf K}}
\def\L{{\bf L}}
\def\M{{\bf M}}
\def\N{{\bf N}}
\def\O{{\bf O}}
\def\P{{\bf P}}
\def\Q{{\bf Q}}
\def\R{{\bf R}}
\def\S{{\bf S}}
\def\T{{\bf T}}
\def\U{{\bf U}}
\def\V{{\bf V}}
\def\W{{\bf W}}
\def\X{{\bf X}}
\def\Y{{\bf Y}}
\def\Z{{\bf Z}}

\def\a{{\bf a}}
\def\b{{\bf b}}
\def\c{{\bf c}}
\def\d{{\bf d}}
\def\e{{\bf e}}
\def\f{{\bf f}}
\def\g{{\bf g}}
\def\h{{\bf h}}
\def\i{{\bf i}}
\def\j{{\bf j}}
\def\k{{\bf k}}
\def\l{{\bf l}}
\def\m{{\bf m}}
\def\n{{\bf n}}
\def\o{{\bf o}}
\def\p{{\bf p}}
\def\q{{\bf q}}
\def\r{{\bf r}}
\def\s{{\bf s}}
\def\t{{\bf t}}
\def\u{{\bf u}}
\def\v{{\bf v}}
\def\w{{\bf w}}
\def\x{{\bf x}}
\def\y{{\bf y}}
\def\z{{\bf z}}

\def\balpha{\mbox{\boldmath{$\alpha$}}}
\def\bbeta{\mbox{\boldmath{$\beta$}}}
\def\bdelta{\mbox{\boldmath{$\delta$}}}
\def\bgamma{\mbox{\boldmath{$\gamma$}}}
\def\blambda{\mbox{\boldmath{$\lambda$}}}
\def\bsigma{\mbox{\boldmath{$\sigma$}}}
\def\btheta{\mbox{\boldmath{$\theta$}}}
\def\bomega{\mbox{\boldmath{$\omega$}}}
\def\bxi{\mbox{\boldmath{$\xi$}}}
\def\bnu{\mbox{\boldmath{$\nu$}}}                                  
\def\bphi{\mbox{\boldmath{$\phi$}}}
\def\bmu{\mbox{\boldmath{$\mu$}}}

\def\bDelta{\mbox{\boldmath{$\Delta$}}}
\def\bOmega{\mbox{\boldmath{$\Omega$}}}
\def\bPhi{\mbox{\boldmath{$\Phi$}}}
\def\bLambda{\mbox{\boldmath{$\Lambda$}}}
\def\bSigma{\mbox{\boldmath{$\Sigma$}}}
\def\bGamma{\mbox{\boldmath{$\Gamma$}}}
\def\bbE{\mathbb{$E$}}
\def\bbR{\mathbb{$R$}}                                  

\newcommand{\myminimum}[1]{\mathop{\textrm{minimum}}_{#1}}
\newcommand{\mymaximum}[1]{\mathop{\textrm{maximum}}_{#1}}    
\newcommand{\mymin}[1]{\mathop{\textrm{minimize}}_{#1}}
\newcommand{\mymax}[1]{\mathop{\textrm{maximize}}_{#1}}
\newcommand{\mymins}[1]{\mathop{\textrm{min.}}_{#1}}
\newcommand{\mymaxs}[1]{\mathop{\textrm{max.}}_{#1}}  
\newcommand{\myargmin}[1]{\mathop{\textrm{argmin}}_{#1}} 
\newcommand{\myargmax}[1]{\mathop{\textrm{argmax}}_{#1}} 
\newcommand{\myst}{\textrm{s.t. }}

\newcommand{\denselist}{\itemsep -1pt}
\newcommand{\sparselist}{\itemsep 1pt}

\definecolor{pink}{rgb}{0.9,0.5,0.5}
\definecolor{purple}{rgb}{0.5, 0.4, 0.8}   
\definecolor{gray}{rgb}{0.3, 0.3, 0.3}
\definecolor{mygreen}{rgb}{0.2, 0.6, 0.2}

\newcommand{\cyan}[1]{\textcolor{cyan}{#1}}
\newcommand{\red}[1]{\textcolor{red}{#1}}  
\newcommand{\blue}[1]{\textcolor{blue}{#1}}
\newcommand{\magenta}[1]{\textcolor{magenta}{#1}}
\newcommand{\pink}[1]{\textcolor{pink}{#1}}
\newcommand{\green}[1]{\textcolor{green}{#1}} 
\newcommand{\gray}[1]{\textcolor{gray}{#1}}    
\newcommand{\mygreen}[1]{\textcolor{mygreen}{#1}}    
\newcommand{\purple}[1]{\textcolor{purple}{#1}}       

\definecolor{greena}{rgb}{0.4, 0.5, 0.1}
\newcommand{\greena}[1]{\textcolor{greena}{#1}}

\definecolor{bluea}{rgb}{0, 0.4, 0.6}
\newcommand{\bluea}[1]{\textcolor{bluea}{#1}}
\definecolor{reda}{rgb}{0.6, 0.2, 0.1}
\newcommand{\reda}[1]{\textcolor{reda}{#1}}

\def\changemargin#1#2{\list{}{\rightmargin#2\leftmargin#1}\item[]}
\let\endchangemargin=\endlist
                                               
\newcommand{\cm}[1]{}

\newcommand{\mtodo}[1]{{\color{red}$\blacksquare$\textbf{[TODO: #1]}}}
\newcommand{\myheading}[1]{\vspace{1ex}\noindent \textbf{#1}}
\newcommand{\htimesw}[2]{\mbox{$#1$$\times$$#2$}}
\newcommand{\mh}[1]{\textcolor{magenta}{[Minh: {#1}]}}
\newcommand{\ms}[1]{\textcolor{red}{[MS: {#1}]}}

\newif\ifshowsolution
\showsolutiontrue

\ifshowsolution  
\newcommand{\Comment}[1]{\paragraph{\bf $\bigstar $ COMMENT:} {\sf #1} \bigskip}
\newcommand{\Solution}[2]{\paragraph{\bf $\bigstar $ SOLUTION:} {\sf #2} }
\newcommand{\Mistake}[2]{\paragraph{\bf $\blacksquare$ COMMON MISTAKE #1:} {\sf #2} \bigskip}
\else
\newcommand{\Solution}[2]{\vspace{#1}}
\fi

\newcommand{\truefalse}{
\begin{enumerate}
	\item True
	\item False
\end{enumerate}
}

\newcommand{\yesno}{
\begin{enumerate}
	\item Yes
	\item No
\end{enumerate}
}
\newcommand{\Sref}[1]{Sec.~\ref{#1}}
\newcommand{\Eref}[1]{Eq.~(\ref{#1})}
\newcommand{\Fref}[1]{Fig.~\ref{#1}}
\newcommand{\Tref}[1]{Tab.~\ref{#1}}

\definecolor{customgray}{rgb}{0.9, 0.9, 0.9}
\newcolumntype{d}{>{\columncolor[gray]{.7}}c}
\newcolumntype{g}{>{\columncolor[gray]{.9}}c}
\newcolumntype{z}{>{\columncolor[gray]{.9}}l}

\maketitle

\begin{abstract}
   In crowd counting, each training image contains multiple people, where each person is annotated by a dot. Existing crowd counting methods need to use a Gaussian to smooth each annotated dot or to estimate the likelihood of every pixel given the annotated point. In this paper, we show that imposing Gaussians to annotations hurts generalization performance. Instead, we propose to use Distribution Matching for crowd COUNTing (DM-Count). In DM-Count, we use Optimal Transport (OT) to measure the similarity between the normalized predicted density map and the normalized ground truth density map. To stabilize OT computation, we include a Total Variation loss in our model. We show that the generalization error bound of DM-Count is tighter than that of the Gaussian smoothed methods. In terms of Mean Absolute Error, DM-Count outperforms the previous state-of-the-art methods by a large margin on two large-scale counting datasets, UCF-QNRF and NWPU, and achieves the state-of-the-art results on the ShanghaiTech and UCF-CC50 datasets. DM-Count reduced the error of the state-of-the-art published result by approximately 16\%. 
   
   Code is available at \url{https://github.com/cvlab-stonybrook/DM-Count}.
\end{abstract}

\section{Introduction}

Image-based crowd counting is an important research problem with various applications in many domains including journalism and surveillance. Current state-of-the-art methods~\cite{Xu_2019_ICCV,Cheng_2019_ICCV,Liu_2019_ICCV,Yan_2019_ICCV,Zhao_2019_CVPR,Zhang_2019_CVPR,Jiang_2019_CVPR,Wan_2019_CVPR,Lian_2019_CVPR,Liu_2019_CVPR, m_Ranjan-etal-ACCV20} treat crowd counting as a density map estimation problem, where a deep neural network first produces a 2D crowd density map for a given input image and subsequently estimates the total size of the crowd by summing the density values across all spatial locations of the density map. For images of large crowds, this density map estimation approach has been shown to be more robust than the detection-then-counting approach~\cite{lin2001estimation, li2008estimating, zhao2003bayesian, ge2009marked} because the former is less sensitive to occlusion and it does not need to commit to binarized decisions at an early stage.

A crucial step in the development of a  density map estimation method is the training  of a deep neural network that maps from an input image to the corresponding annotated density map. In all existing crowd counting datasets~\cite{idrees2018composition, zhang2016single, idrees2013multi, wang2020nwpu}, the annotated density map for each training image is a sparse binary mask, where each individual person is marked with a single dot on their head or forehead. The spatial extent of each person is not provided, due to the laborious effort needed for delineating the spatial extent, especially when there is too much occlusion ambiguity. Given training images with dot annotation, training the density map estimation network is equivalent to optimizing the parameters of the network to minimize a differentiable loss function that measures the discrepancy between the predicted density map and the dot-annotation map. Notably, the former is a dense real-value matrix, while the later is a sparse binary matrix. Given the sparsity of the dots, a function that is defined based on the pixel-wise difference between the annotated and predicted density maps is hard to train because the reconstruction loss is heavily unbalanced between the 0s and 1s in the sparse binary matrix. One approach to alleviate this problem is to turn each annotated dot into a Gaussian blob such that the ground truth is more balanced and thus the network is easier to train. Almost all prior crowd density map estimation methods~\cite{zhang2019relational, zhang2019attentional, zhang2016single,sam2017switching,li2018csrnet,onoro2016towards,ranjan2018iterative,cao2018scale,liu2018leveraging,wang2019learning,liu2019context,shi2019revisiting,liu2019point,liu2019adcrowdnet} have followed this convention. 
Unfortunately, the performance of the resulting network is highly dependent on the quality of this ``pseudo ground truth'', but it is not trivial to set the right widths for the Gaussian blobs given huge variation in the sizes and shapes of people in a perspective image of a crowded scene.

Recently,~\citet{bayesianCounting} proposed a Bayesian loss to measure the discrepancy between the predicted and the annotated density maps. This method transforms a binary ground truth annotation map into $N$ ``smoothed ground truth'' density maps, where $N$ is the count number. Each pixel value of a smoothed ground truth density map is the posterior probability of the corresponding annotation dot given the location of that pixel. 
Empirically, this method has been shown to outperform other aforementioned approaches~\cite{ zhang2016single,sam2017switching,li2018csrnet,onoro2016towards,ranjan2018iterative,cao2018scale}. However, there are two major problems with this loss function. First, it also requires a Gaussian kernel to construct the likelihood function for each annotated dot, which involves setting the kernel width. Second, 
this loss corresponds to an underdetermined system of equations with infinitely many solutions. The loss can be 0 for many density maps that are not similar to the ground truth density map. As a consequence, using this loss for training can lead to a predicted density map that is very different from the ground truth density map. 


In this paper, we address the shortcomings in existing approaches with the following contributions.

\begin{itemize} [leftmargin=3.5mm,noitemsep,topsep=0pt,parsep=0pt,partopsep=0pt]
\item
We theoretically and empirically  show that imposing Gaussians to annotations will hurt the generalization performance of a crowd counting network.
\item We propose DM-Count, a method that performs Distribution Matching for crowd COUNTing. Unlike previous works, DM-Count does not need any Gaussian smoothing ground truth annotations. Instead, we use Optimal Transport (OT) to measure the similarity between the normalized predicted density map and the normalized ground truth density map. To stabilize the OT computation, we further add a Total Variation (TV) loss.
\item
We present the generalization error bounds for the counting loss, OT loss, TV loss and the overall loss in our method. All the bounds are tighter than those of the Gaussian smoothed methods.  
\item
Empirically, our method improved the state-of-the-art  by a large margin on four challenging crowd counting datasets: UCF-QNRF, NWPU, ShanghaiTech, and UCF-CC50.  
Notably, our method reduced the published state-of-the-art MAE on the NWPU dataset by approximately 16\%. 
\end{itemize}

\section{Previous Work}

\subsection{Crowd Counting Methods}
Crowd counting methods can be divided into three categories: detection-then-count, direct count regression, and density map estimation. 
Early methods~\cite{lin2001estimation, li2008estimating, zhao2003bayesian, ge2009marked}  detect people, heads, or upper bodies in the image. However, accurate detection is difficult for dense crowds. Besides, it also requires bounding box annotation, which is a laborious and ambiguous process due to heavy occlusion. Later methods~\cite{chan2009bayesian,chen2012feature, wang2015deep, chen2013cumulative} avoid the detection problem and directly learn to regress the count from a feature vector. But their results are less interpretable and the dot annotation maps are underutilized. 
Most recent works~\cite{ li2018csrnet,ranjan2018iterative,idrees2018composition,cao2018scale,bayesianCounting,wang2019learning,liu2019context,shi2019revisiting,liu2019point,liu2019adcrowdnet,Xu_2019_ICCV,Cheng_2019_ICCV,Liu_2019_ICCV,Yan_2019_ICCV,Zhao_2019_CVPR,Zhang_2019_CVPR,Jiang_2019_CVPR,Wan_2019_CVPR,Lian_2019_CVPR,Liu_2019_CVPR,lu2018class,wan2019adaptive,xiong2019open, sindagi2019multi,sam2018top,zhang2019relational,laradji2018blobs,shen2018crowd,liu2018decidenet,tian2019padnet,sindagi2019ha} are based on density map estimation, which has been shown to be more robust than detection-then-count and count regression approaches. 

Density map estimation methods usually define the training loss based on the pixel-wise difference between the Gaussian smoothed density map and the predicted density map. Instead of using a single kernel width to smooth the dot annotation, \cite{zhang2016single, idrees2013multi, wan2019adaptive} used adaptive kernel width. The kernel width is selected based on the distance to an annotated dot's nearest neighbors. Specifically, \cite{idrees2018composition} generated multiple smoothed ground truth density maps on different density levels. The final loss combines the reconstruction errors from multiple density levels. However, these methods assume the crowd is evenly distributed; in reality crowd distribution is quite irregular. The Bayesian loss method~\cite{bayesianCounting} uses a Gaussian to construct a likelihood function for each annotated dot. However, it may not predict a correct density because the loss is underdetermined. Detailed analysis can be found in Sec~\ref{sec:bayesian}.

\subsection{Optimal Transport}


We propose a novel loss function based on Optimal Transport (OT)~\cite{villani2008optimal}. For a better understanding of the proposed method, we briefly review the Monge-Kantarovich OT formulation in this section. 

Optimal Transport refers to the optimal cost to transform one probability distribution to another. Let $\mX = \{ \x_i | \x_i \in \mathbb{R}^d\}_{i=1}^n$ and $\mY = \{ \y_j | \y_j \in \mathbb{R}^d \}_{j=1}^n$ be two sets of points on $d$-dimensional vector space. Let $\bmu$ and $\bnu$ be two probability measures defined on $\mX$ and $\mY$, respectively; $\bmu, \bnu \in \mathbb{R}^n_{+}$ and $\1_n^T\bmu = \1_n^T\bnu = 1$ ($\1_n$ is a $n$-dimensional vector of all ones). Let $c: \mX \times \mY \mapsto \mathbb{R}_{+}$ be the cost function for moving from a point in $\mX$ to a point in $\mY$, and $\C$ be the corresponding $n{\times}n$  cost matrix for the two sets of points: $\C_{ij} = c(\x_i, \y_j)$. Let $\Gamma$ be the set of all possible ways to transport probability mass from $\mX$ to $\mY$: $\Gamma = \{ \gamma \in \mathbb{R}^{n \times n}_{+}: \gamma \boldsymbol{1} = \bmu, \gamma^{T} \boldsymbol{1} = \bnu \}$. The Monge-Kantorovich's Optimal Transport (OT) cost between $\bmu$ and $\bnu$ is defined as: 
\begin{align}
\mW(\bmu, \bnu) = \min_{\gamma \in \Gamma} ~~& \langle \C, \gamma \rangle. 
\end{align}
Intuitively, if the probability distribution $\bmu$ is viewed as a unit amount of ``dirt'' piled on $\mX$ and $\bnu$ a unit amount of dirt piled on $\mY$, the OT cost is the minimum ``cost'' of turning one pile into the other. The OT cost is a principal measurement to quantify the dissimilarity between two probability distributions, also taking into account the distance between ``dirt'' locations. 

The OT cost can also be computed via the dual formulation: 
    \begin{align}
    \label{eq: ot_dual}
\mW(\bmu, \bnu) = \max_{\balpha, \bbeta \in \mathbb{R}^{n}} & \langle \balpha, \bmu \rangle + \langle \bbeta, \bnu \rangle, \quad  \myst  \alpha_i + \beta_j \leq c(\x_i, \y_j),~\forall i, j.
    \end{align}

\section{DM-Count: Distribution Matching for Crowd Counting}



We consider crowd counting as a distribution matching problem. In this section, we propose DM-Count: Distribution matching for crowd counting. 
A network for crowd counting inputs an image and outputs a map of density values. The final count estimate can be obtained by summing over the predicted density map. DM-Count is agnostic to different network architectures. In our experiments, we use the same network as in the Bayesian loss paper~\cite{bayesianCounting}. Unlike all previous density map estimation methods which need to use Gaussians to smooth ground truth annotations, DM-Count does not need any Gaussian to preprocess ground truth annotations. 



Let $\z \in \mathbb{R}^n_{+}$ denote the vectorized binary map for dot-annotation and $\hat{\z} \in \mathbb{R}^n_{+}$ the vectorized predicted density map returned by a neural network. By viewing $\z$ and $\hat{\z}$ as unnormalized density functions, we formulate the loss function in DM-Count using three terms: the counting loss, the OT loss, and the Total Variation (TV) loss. The first term measures the difference between the total masses, while the last two measures the difference between the distributions of the normalized density functions. 

\myheading{The Counting Loss}. Let $\normone{\cdot}$ denote the $L_1$ norm of a vector, and so $\normone{\z}$, $\normone{\hat{\z}}$ are the ground truth and predicted counts respectively. The goal of crowd counting is to make $\normone{\hat{\z}}$ as close as possible to $\normone{\z}$, and the counting loss is defined as the absolute difference between them: 
\begin{align}
\label{eq: counting_loss}
    \ell_C \left  (\z, \hat{\z} \right) = | \normone{\z} - \normone{\hat{\z}} |. 
\end{align}
\myheading{The Optimal Transport Loss}.
Both $\z$ and $\hat{\z}$ are unnormalized density functions, but we can turn them into probability density functions (pdfs) by dividing them by the their respective total mass. Apart from OT, the Kullback-Leibler divergence and Jensen-Shannon divergence can also measure the similarity between two pdfs. However, these measurements do not provide valid gradients to train a network if the source distribution does not overlap with the target distribution \cite{martin2017wasserstein}. Therefore, we propose the use of OT in this work. We define the OT loss as follows: 
\begin{align}
\ell_{OT} \left(\z, \hat{\z} \right ) = \mW\left( \frac{\z}{\normone{\z}}, \frac{\hat{\z}}{\normone{\hat{\z}}}\right ) = \left\langle \balpha^*, \frac{\z}{\normone{\z}}  \right\rangle + \left\langle \bbeta^*, \frac{\hat{\z}}{\normone{\hat{\z}}} \right\rangle, \label{eqn:OT4z}
\end{align}
where $\balpha^*$ and $\bbeta^*$ are the solutions of Problem (\ref{eq: ot_dual}). We use the quadratic transport cost, i.e., $c \left (\z(i), \hat{\z}(j) \right) = \norm{\z(i) - \hat{\z}(j)}_2^2$, where $\z(i)$ and $\hat{\z}(j)$ are 2D coordinates of locations $i$ and $j$, respectively. To avoid the division-by-zero error, we add a machine precision to the denominator. 

Since the entries in $\hat{\z}$ are non-negative, the gradient of Eq.~(\ref{eqn:OT4z}) with respect to  $\hat{\z}$ is:
\begin{align}
\label{eq: ot_grad}
	\frac{\partial \ell_{OT}\left(\z, \hat{\z} \right )  }{\partial \hat{\z}} = \frac{\bbeta^*}{\normone{\hat{\z}}} - \frac{ \langle \bbeta^*, \hat{\z} \rangle }{\normone{\hat{\z}}^2}.
\end{align}
This gradient can be back-propagated to learn the parameters of the density estimation network.


\myheading{Total Variation Loss}. In each training iteration, we use the Sinkhorn algorithm \cite{peyre2019computational} to approximate $\balpha^*$ and $\bbeta^*$. The time complexity is $O(n^2 \log n  / \epsilon^2)$ \cite{pmlr-v80-dvurechensky18a}, where $\epsilon$ is the desired optimality gap, i.e., the upper bound for the difference between the returned objective and the optimal objective. When optimizing with the Sinkhorn algorithm, the objective decreases dramatically at the beginning but only converges slowly to the optimal objective in later iterations. In practice, we set the maximum number of iterations, and the Sinkhorn algorithm only returns an approximate solution. As a result, when we optimize the OT loss with the Sinkhorn algorithm, the predicted density map  ends up close to the ground truth density map, but not exactly the same. The OT loss will approximate well the dense areas of the crowd, but the approximation might be poorer for the low density areas of the crowd. To address this issue, we  additionally use the Total Variation (TV) loss, defined as\footnote{In the  training loss context, Total Variation refers to the total variation distance of two
probability measures. A formal definition can be found in  \cite[Definition 2.4, page 83]{tsybakov2008introduction}. Eq. (\ref{eq: tvloss}) is \cite[Lemma 2.1, page 84]{tsybakov2008introduction}.}: 
\begin{align}
\label{eq: tvloss}
\ell_{TV}(\z, \hat{\z}) = \left\Vert \frac{\z}{\normone{\z}} - \frac{\hat{\z}}{\normone{\hat{\z}}}\right\Vert_{TV} = \frac{1}{2} \left\Vert \frac{\z}{\normone{\z}} - \frac{\hat{\z}}{\normone{\hat{\z}}}\right\Vert_1.
\end{align}
The TV loss will also increase the stability of the training procedure. Optimizing the OT loss with the Sinkhorn algorithm is a min-max saddle point optimization procedure, which is similar to GAN optimization~\cite{goodfellow2014generative}. The stability of GAN training can be increased by adding a reconstruction loss, as shown in the Pix2Pix GAN~\cite{isola2017image}. To this end, the TV loss is similar to the reconstruction loss, and also increases the stability of the training procedure.

The gradient of the TV loss with respect to the predicted density map $\hat{\z}$ is: 
\begin{align}
\label{eq: tv_grad}
	\frac{\partial \ell_{TV}\left(\z, \hat{\z} \right )  }{\partial \hat{\z}} = - \frac{1}{2}\left (  \frac{\mbox{sign}(\v)}{\normone{\hat{\z}}} - \frac{ \langle \mbox{sign} (\v), \hat{\z} \rangle }{\normone{\hat{\z}}^2} \right),
\end{align}
where $\v = \z / \normone{\z} - \hat{\z} / \normone{\hat{\z}}$, and $\mbox{sign}(\cdot)$ is the Sign function on each element of a vector.

\myheading{The Overall Objective}. The overall loss function is the combination of the counting loss, the OT loss, and the TV loss:
\begin{align}
\ell(\z, \hat{\z}) =  \ell_{C} (\z, \hat{\z}) + \lambda_1 \ell_{OT}( \z, \hat{\z}) + \lambda_2  \normone{\z} \ell_{TV}(\z, \hat{\z}) ,
\end{align}
where $\lambda_1$ and $\lambda_2$ are tunable hyper-parameters for the OT and TV losses. To ensure that the TV loss has the same scale as the counting loss, we multiply this loss term with the total count.

 
Given $K$ training images $\{I_k\}_{k=1}^K$ with corresponding dot annotation maps $\{\z_k\}_{k=1}^K$, we will learn a deep neural network $f$ for density map estimation by minimizing:
$L(f) = \frac{1}{K}\sum_{k=1}^{K} \ell(\z_k, f(I_k))$.

\section{Generalization Bounds and Theoretical Analysis}
In this section, we analyze the theoretical properties of the Gaussian smoothed methods, the Bayesian loss, and the proposed DM-Count. The proofs of the theorems in this section can be found in the supplementary material. First, we introduce some notations below.

Let $\mI$ denote the set of images and $\mZ$ the set of dot annotation maps. Let $\mD = \{(I, \z)\} $ be the joint distribution of crowd images and corresponding dot annotation maps. Let $\mH$ be a hypothesis space. Each $h \in \mH$ maps from $I \in \mI$ to each dimension of $\z \in \mZ$. Let $\mathcal{F} = \mathcal{H} \times \dots \times \mathcal{H}$ ($n$ times) be the mapping space. Each $f \in  \mF$ maps $I \in \mI$ to $\z \in \mZ$. Let $\t$ be the Gaussian smoothed density map of each $\z \in \mD$, and let $\Tilde{\mD} = \{ (I, \t) \}$ be the joint distribution of $(I, \t)$. Let $S = \{(I_k, \z_k)\}_{k=1}^{K}$, and $\Tilde{S} = \{(I_k, \t_k)\}_{k=1}^{K}$ be the finite sets of $K$ samples i.i.d. sampled from $\mD$ and $\Tilde{\mD}$, respectively. Let $R_S(\mathcal{H})$ denote the empirical Rademacher complexity~\cite{bartlett2002rademacher} for $\mathcal{H}$ w.r.t $S$.  Given a data set $D \in \{ \mD, S, \Tilde{\mD}, \Tilde{S} \}$, a mapping $f \in \mF$ and a loss function $\ell$, let $\mR(D, f, \ell) = \mathbb{E}_{(I, \s) \sim D} [ \ell(\s, f(I))]$ denote the expected risk. Let $\ell_1(\z, \hat{\z}) = \normone{\z - \hat{\z}}$. Let $f^{D}_{\Delta} = \myargmin{f \in \mF} \mR (D, f, \ell_{\Delta})$ be the minimizer of $\mR (D, f, \ell_{\Delta})$ over a data set $D$ using the loss $\ell_{\Delta}$, where $D \in \{ \mD, S, \Tilde{\mD}, \Tilde{S} \}$, and $\Delta \in \{1, C, OT, TV, \emptyset \}$.

\subsection{Generalization Error Bounds of Gaussian Smoothed Methods}
Many existing methods (e.g.,~\cite{ zhang2016single,li2018csrnet,ranjan2018iterative}) use Gaussian-smoothed annotation maps for training. Below we give  generalization error bounds when using the $\ell_1$ loss on the density maps.
\begin{theorem}
\label{thm: gs_bound}
Assume that $\forall f \in \mathcal{F}$ and $(I, \t) \sim \Tilde{\mathcal{D}}$, we have $\ell(\t, f(I)) \leq B$. Then, for any $0 < \delta < 1$, with probability of at least $1 - \delta$, \\
a) the upper bound of the generalization error is
\begin{align}
    \mR(\mD, f^{\Tilde{S}}_1, \ell_1) \leq \mR(\Tilde{\mD}, f^{\Tilde{\mD}}_1, \ell_1) + 2 n R_S(\mathcal{H}) 
    + 5 B \sqrt{2\log{(8/\delta)}/K} + \mathbb{E}_{(I, \z) \sim \mathcal{D}} \normone{\z - \t}, \nonumber 
\end{align}
b) the lower bound of the generalization error is
\begin{align}
    \mR(\mD, f^{\Tilde{S}}_1, \ell_1) \geq \left | \mathbb{E}_{(I, \z) \sim \mathcal{D}}  \normone{\z - \t} - \mR(\Tilde{\mD}, f^{\Tilde{S}}_1, \ell_1) \right|. \nonumber 
\end{align}
\end{theorem}


In this theorem, as the number of samples $K$ grows to infinity, $2 n R_S(\mathcal{H})$ and $5 B \sqrt{2\log{(8/\delta)}/K}$ decrease to 0. Theorem \ref{thm: gs_bound}.a) shows that the upper bound (worst case) of the expected risk $\mR(\mD, f^{\Tilde{S}}_1, \ell_1)$, which is evaluated on real ground truth data using an empirical minimizer trained on the Gaussian smoothed ground truth, does not exceed $\mR(\Tilde{\mD}, f^{\Tilde{\mD}}_1, \ell_1) + \mathbb{E}_{(I, \z) \sim \mathcal{D}} \normone{\z - \t}$ given sufficient training data. Theorem \ref{thm: gs_bound}.b) shows 
that the lower bound (best case) of $\mR(\mD, f^{\Tilde{S}}_1, \ell_1)$ is not smaller than $ | \mathbb{E}_{(I, \z) \sim \mathcal{D}}  \normone{\z - \t} - \mR(\Tilde{\mD}, f^{\Tilde{S}}_1, \ell_1)| $. This means that if $\mR(\Tilde{\mD}, f^{\Tilde{S}}_1, \ell_1) \leq \mathbb{E}_{(I, \z) \sim \mathcal{D}}  \normone{\z - \t}$, then the smaller  $\mR(\Tilde{\mD}, f^{\Tilde{S}}_1, \ell_1)$ is, the larger the expected risk $\mR(\mD, f^{\Tilde{S}}_1, \ell_1)$ will be. In other words, the better a good model $f^{\Tilde{S}}_1$ performs on the Gaussian smoothed ground truth $\Tilde{\mD}$, the poorer it generalizes on the real ground truth $\mD$. Furthermore, as long as $\mR(\Tilde{\mD}, f^{\Tilde{S}}_1, \ell_1) \ne \mathbb{E}_{(I, \z) \sim \mathcal{D}}  \normone{\z - \t}$, we have $\mR(\mD, f^{\Tilde{S}}_1, \ell_1) > 0$. $\mR(\mD, f^{\Tilde{S}}_1, \ell_1)$ can be as large as $\mathbb{E}_{(I, \z) \sim \mathcal{D}} \normone{\z - \t}$ when $\mR(\Tilde{\mD}, f^{\Tilde{S}}_1, \ell_1) = 0$. This is undesirable because we want the risk $\mR(\mD, f^{\Tilde{S}}_1, \ell_1)$ evaluated on the real ground truth to be 0 as well. 






\subsection{The Underdetermined Bayesian Loss}\label{sec:bayesian}
The Bayesian Loss~\cite{bayesianCounting} is: 
\begin{align}
    \ell_{Bayesian}(\z, \hat{\z}) = \sum_{i=1}^{N} \lvert 1 - \langle \p_i, \hat{\z} \rangle \rvert, ~\textrm{where} \quad  \p_i = \frac{\mN (\q_i, \sigma^2 \1_{2\times2})}{\sum_{i=1}^{N} \mN (\q_i, \sigma^2 \1_{2\times2})},
\end{align}
and $N$ is number of people of $\z$, and $\mN (\q_i, \sigma^2 \1_{2\times2})$ is a Gaussian distribution centered at $\q_i$ with variance $\sigma^2\1_{2\times2}$. $\q_i$ is the $i^{th}$ annotated dot in $\z$. The dimension of $\p_i$ and $\z$ is $n$, the number of pixels of the density map. However, since the number of annotated dots $N$ is less than $n$, the Bayesian loss is underdetermined. For a ground truth annotation $\z$, there are infinitely many $\hat{\z}$ with  $\ell_{Bayesian}(\z, \hat{\z}) = 0$ and $\hat{\z} \ne \z$. Therefore, the predicted density map could be very different from the ground truth density map.


\subsection{The Generalization Error Bounds of the Losses in DM-Count}
We give the generalization error bounds of the losses in the proposed method in the following theorem. 
\begin{theorem}
\label{thm: proposed_bounds}
Assume that $\forall f \in \mathcal{F}$ and $(I, \z) \sim \mathcal{D}$, we have $\normone{\z} \ge 1$, $\normone{f(I)} \ge 1$ (can be satisfied by adding a dummy dimension with value of 1 to both $\z$ and $f(I)$) and $\ell_C(\z, f(I)) \leq B$. Then, for any $0 < \delta < 1$, with probability of at least $1 - \delta$\\
a) the generalization error bound of the counting loss is
\begin{align}
    \mR(\mD, f^{S}_C, \ell_C) \leq \mR(\mD, f^{\mD}_C, \ell_C)  + 2 n R_S(\mathcal{H}) + 5 B \sqrt{2\log{(8/\delta)}/K} \nonumber,
\end{align}
b) the generalization error bound of the OT loss is 
\begin{align}
    \mR(\mD, f^{S}_{OT}, \ell_{OT}) \leq \mR(\mD, f^{\mD}_{OT}, \ell_{OT}) + 4 \C_{\infty} n^2 R_S(\mathcal{H}) + 5\C_{\infty}\sqrt{2\log{(8/\delta)}/K} \nonumber,
\end{align}
c) the generalization error bound of the TV loss is 
\begin{align}
     \mR(\mD, f^{S}_{TV}, \ell_{TV}) \leq \mR(\mD, f^{\mD}_{TV}, \ell_{TV}) + n^2 R_S(\mathcal{H}) + 5 \sqrt{2\log{(8/\delta)}K} \nonumber,
\end{align}
d) the generalization error bound of the overall loss is 
\begin{align}
\nonumber
    \mR(\mD, f^{S}, \ell) \leq \mR(\mD, f^{\mD}, \ell)
     + (2n + 4 \lambda_1 \C_{\infty} n^2 + \lambda_2 N n^2) R_S(\mathcal{H}) \\\nonumber
    + 5(B + \lambda_1 \C_{\infty} + \lambda_2 N )\sqrt{2\log{(8/\delta)}K} \nonumber,
\end{align}
where $\C_{\infty}$ is the maximum cost in the cost matrix in OT, and $N = \sup \{ \normone{\z} ~|~ \forall (I, \z) \sim \mathcal{D} \}$ is the maximum count number over a dataset.
\end{theorem}

In the above theorem, as $K$ grows, $R_S(\mH)$ and $\sqrt{2\log{(1/\delta)}K}$ decrease. All the expected risks $\mR(\mD, f^S_{\Delta}, \ell_{\Delta})$ using the empirical minimizers $f^S_{\Delta}$ converge to the expected risks $\mR(\mD, f^{\mD}_{\Delta}, \ell_{\Delta})$, $\Delta \in \{ C, OT, TV, \emptyset \}$ using optimal minimizers $f^{\mD}_{\Delta}$. This means that all the upper bounds are tight. In addition, all upper bounds are tighter than the upper bound of the Gaussian smoothed methods shown in Theorem \ref{thm: gs_bound}.a). The bound of the OT loss in Theorem \ref{thm: proposed_bounds}.b) is related to the maximum transport cost $\C_{\infty}$. Therefore, we need to use a smaller transport cost in OT for better generalization performance. The coefficient of $R_S(\mH)$ for the counting loss is $O(n)$, and for the OT loss and the TV loss is $O(n^2)$. This means that for larger image size, we need more images to train. The number is linear to the size of $\z$ using solely the counting loss, and quadratic using solely the OT loss or the TV loss. When using all three losses, we need to set $\lambda_1$ and $\lambda_2$ to be small in order to balance the three losses.

\section{Experiments}
In this section, we describe experiments on toy data and on benchmark crowd counting datasets. More detailed dataset descriptions, implementation details and experimental settings can be found in the supplementary material. 

\def\subFigSz{0.24\linewidth} 
\begin{figure*}[!t]
\makebox[\subFigSz]{\bf Target density map}
\makebox[\subFigSz]{\bf  Pixel-wise loss}
\makebox[\subFigSz]{\bf  Bayesian loss}
\makebox[\subFigSz]{\bf  DM-Count (proposed)}

\begin{subfigure}[b]{0.24\textwidth}
\centering
	 \includegraphics[width=\textwidth]{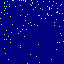}
	 \mbox{\small Count: 115} \\
	 \mbox{}
\end{subfigure}
\begin{subfigure}[b]{0.24\textwidth}
\centering
	 \includegraphics[width=\textwidth]{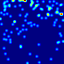}
	 \mbox{\small Count: 113.13} \\
	 \mbox{\small PSNR: 32.91, SSIM: 0.55}
\end{subfigure}
\begin{subfigure}[b]{0.24\textwidth}
\centering
	 \includegraphics[width=\textwidth]{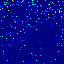}
	 \mbox{\small Count: 114.75} \\
	 \mbox{\small PSNR: 38.61, SSIM: 0.84}
\end{subfigure}
\begin{subfigure}[b]{0.24\textwidth}
\centering
	 \includegraphics[width=\textwidth]{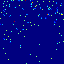}
	 \mbox{\small Count: 114.98} \\
	 \mbox{\small PSNR: 41.84, SSIM: 0.95}
\end{subfigure}     
     
\caption{{\bf Comparison of different methods on toy data.} The pixel-wise loss generates a blurry density map with a higher counting error. The Bayesian loss produces  dissimilar density maps from the ground truth, with high values in many locations with no annotations. DM-Count is able to produce more accurate crowd count and localization than the other two methods. 
\label{fig:toy}}
\end{figure*}

\subsection{Results on Toy Data}

To understand the empirical behavior of different methods, we consider a toy problem where the task is to move a source density map $\hat{\z}$ to a target density map $\z$ using the Pixel-wise loss, the Bayesian loss and DM-Count. 
The source density map $\hat{\z}$ is initialized from a uniform distribution between 0 and 0.01, and the target density map is shown in the leftmost figure in Fig.~\ref{fig:toy}. All three methods start from the same source density map. Fig.~\ref{fig:toy} visualizes the final $\hat{\z}$ at convergence. The Pixel-wise loss yields a blurry density map with a higher count. The Bayesian loss performs better than the Pixel-wise loss in terms of counting error, Peak Signal-to-Noise Ratio (PSNR) and Structural Similarity in Image (SSIM)~\cite{wang2004image}, but the resulting density map is quite different from the target, with high values at many locations where no dots are annotated. This confirms our analysis that the Bayesian loss corresponds to an underdetermined system such that the output density map could be very different from the target density map. In contrast, DM-Count is able to produce a more accurate count and density map. DM-Count outperforms the Bayesian loss by a large margin in both PSNR and SSIM.

\subsection{Results on Benchmark Datasets}
We perform experiments on four challenging crowd counting datasets: UCF-QNRF~\cite{idrees2018composition}, NWPU~\cite{wang2020nwpu}, ShanghaiTech~\cite{zhang2016single}, and UCF-CC-50~\cite{idrees2013multi}. It is worth noting that the \textbf{NWPU} dataset is the largest-scale and most challenging crowd counting dataset publicly available today. The ground truth counts for test images are not released, and the results on the test set must be obtained by submitting to the evaluation server at \url{https://www.crowdbenchmark.com/nwpucrowd.html}. Following previous work~\cite{ranjan2018iterative,idrees2018composition,cao2018scale,idrees2013multi,zhang2016single}, we use the following metrics: Mean Absolute Error (MAE), Root Mean Squared Error (RMSE), and Mean Normalized Absolute Error (NAE) as evaluation metrics. For all three metrics, the smaller the better. For a fair comparison, we use the same network as in the Bayesian loss paper~\cite{bayesianCounting}. In all experiments, we set $\lambda_1=0.1, \lambda_2 = 0.01$, and the Sinkhorn entropic regularization parameter  to 10.
The number of Sinkhorn iterations is set to 100. On average, the OT computation time is 25\textit{ms} for each image.

\begin{table}[!tb]
\centering
\begin{tabular}{lcc|cc|cc|cc}
\toprule
         & \multicolumn{2}{c}{UCF-QNRF} & \multicolumn{2}{c}{ShanghaiTech A}  & \multicolumn{2}{c}{ShanghaiTech B}  & \multicolumn{2}{c}{UCF-CC-50} \\
               & MAE          & RMSE & MAE          & RMSE & MAE          & RMSE & MAE          & RMSE  \\
\midrule
Crowd CNN~\cite{zhang2015cross} &-&-  &        181.8      &  277.7           &   32.0           &  49.8       & 467.0 & 498.5    \\
MCNN~\cite{zhang2016single} & 277 & 426 &     110.2     &    173.2         &  26.4            &  41.3 & 377.6 & 509.1  \\
CMTL~\cite{sindagi2017cnn}& 252 & 514 & 101.3 & 152.4 & 20.0 & 31.1 & 322.8 & 341.4  \\
Switch CNN ~\cite{Sam-etal-CVPR17} & 228 & 445  & 90.4           &  135.0           &     21.6         &     33.4   & 318.1& 439.2      \\
IG-CNN~\cite{babu2018divide} &-&- & 72.5 & 118.2 & 13.6 & 21.1  & 291.4 & 349.4 \\
ic-CNN~\cite{ranjan2018iterative}  & - & - & 68.5 &116.2 & 10.7& 16.0  &260.9 & 365.5\\
CSR Net~\cite{li2018csrnet} & -&- & 68.2 & 115.0 & 10.6 & 16.0  &  266.1 & 397.5\\
SANet~\cite{cao2018scale}  & -&-  &  67.0 & 104.5 & 8.4 & 13.6 &  258.4 & 334.9  \\
CL-CNN~\cite{idrees2018composition} & 132 & 191 &-&-&-&-&-&- \\
PACNN ~\cite{shi2019revisiting} & -&-  & 62.4 & 102.0 & 7.6 & 11.8 & 241.7 & 320.7 \\
CAN ~\cite{liu2019context} & 107 & 183 & 62.3 & 100.0 & 7.8 & 12.2  & 212.2 & \textbf{243.7}   \\
SFCN ~\cite{wang2019learning} & 102  &   171 & 64.8  & 107.5  &7.6   &  13.0  & 214.2 & 318.2\\
ANF~\cite{zhang2019attentional} & 110 & 174 & 63.9 & 99.4 &  8.3 & 13.2 &250.2 &340.0  \\
Wan \textit{et al.}~\cite{wan2019adaptive} & 101 & 176& 64.7 & 97.1 & 8.1 & 13.6 &-&- \\
\midrule
Pixel-wise Loss~\cite{bayesianCounting} & 106.8 & 183.7 & 68.6 & 110.1 & 8.5 & 13.9 & 251.6 & 331.3 \\
Bayesian Loss~\cite{bayesianCounting} & 88.7  & 154.8 & 62.8 & 101.8  & 7.7 & 12.7 &229.3 & 308.2 \\
DM-Count (proposed)  & \textbf{85.6} & \textbf{148.3} & \textbf{59.7} & \textbf{95.7} & \textbf{7.4} & \textbf{11.8}   & \textbf{211.0} & 291.5 \\
\bottomrule
\end{tabular}
\vskip 0.05in
\caption{{\bf Results on the UCF-QNRF, Shanghai Tech, and UCF-CC-50 datasets}. \label{tab:qnrf_sh_cc}}
\end{table}

\begin{table}[!tb]
\centering
\begin{tabular}{lcccccccc}
\toprule
         & Backbone  & \multicolumn{2}{c}{Validation set} & \multicolumn{3}{c}{Test set} \\
         \cmidrule(lr){3-4} \cmidrule(lr){5-7} 
             &  &  MAE & RMSE &  MAE   & RMSE  & NAE\\
\midrule
MCNN~\cite{zhang2016single} & FS & 218.5 & 700.6 & 232.5 & 714.6 & 1.063  \\
CSR net~\cite{li2018csrnet} & VGG-16 & 104.8 & 433.4 & 121.3 & 387.8 & 0.604\\
PCC-Net-VGG~\cite{gao2019pcc} & VGG-16 & 100.7 & 573.1 & 112.3 & 457.0 & 0.251 \\
CAN ~\cite{liu2019context} & VGG-16 & 93.5 & 489.9 & 106.3 & \textbf{386.5} & 0.295 \\
SCAR~\cite{gao2019scar} & VGG-16 & 81.5 & 397.9 & 110.0 & 495.3 & 0.288\\
Bayesian Loss~\cite{bayesianCounting} &  VGG-19 & 93.6 & 470.3 & 105.4 & 454.2 & 0.203 \\
SFCN~\cite{wang2019learning} & ResNet-101 & 95.4 & 608.3 & 105.7 & 424.1 & 0.254 \\
\midrule
DM-Count (proposed) &  VGG-19  & \textbf{70.5} & \textbf{357.6} & \textbf{88.4} & 388.6 &  \textbf{0.169}  \\
\bottomrule
\end{tabular}
\vskip 0.05in
\caption{{\bf Results of various methods on the NWPU validation and test sets}. \label{tab:nwpu}}
\end{table}

\begin{figure*}[!t]
\centering
\makebox[\subFigSz]{\bf Image}
\makebox[\subFigSz]{\bf  Pixel-wise loss}
\makebox[\subFigSz]{\bf  Bayesian loss}
\makebox[\subFigSz]{\bf  DM-Count (proposed)}
\begin{subfigure}[b]{0.24\textwidth}
\begin{center}
\end{center}
	 \includegraphics[width=\textwidth]{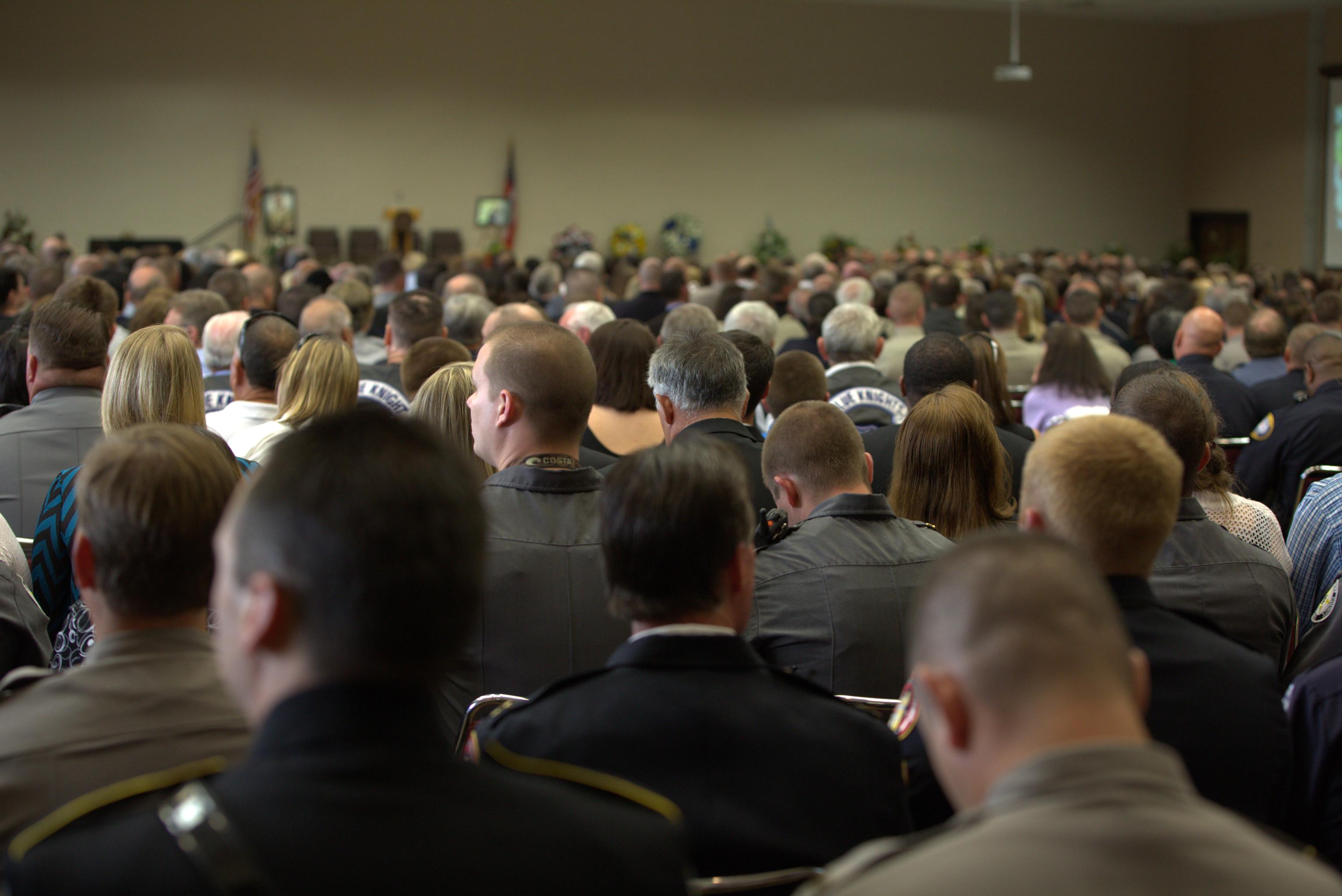}
	 \newlength{\imagewidtha}
     \settowidth{\imagewidtha}{\includegraphics{img_0052.jpg}} 
     \newlength{\imageheighta}
     \settoheight{\imageheighta}{\includegraphics{img_0052.jpg}}
     \centering
	 \includegraphics[trim=0.1\imagewidtha{} 0.55\imageheighta{} 0.7\imagewidtha{} 0.25\imageheighta{}, clip, width=0.485\textwidth]{img_0052.jpg}
	 \includegraphics[trim=0.75\imagewidtha{} 0.6\imageheighta{} 0.05\imagewidtha{} 0.2\imageheighta{}, clip, width=0.485\textwidth]{img_0052.jpg}
	 \mbox{Count: 139}\\
	 \mbox{}
\end{subfigure}
\begin{subfigure}[b]{0.24\textwidth}
\begin{center}
\end{center}
	\includegraphics[width=\textwidth]{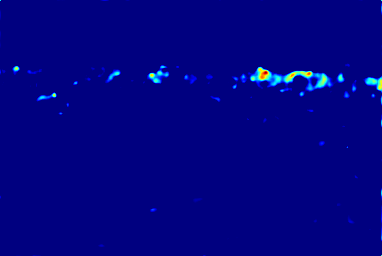}
	\newlength{\imagewidthb}
     \settowidth{\imagewidthb}{\includegraphics{mse_img_0052_mae-105.26_gt-139_pred-33.74_psnr-43.267_ssim-0.834.png}}
     \newlength{\imageheightb}
     \settoheight{\imageheightb}{\includegraphics{mse_img_0052_mae-105.26_gt-139_pred-33.74_psnr-43.267_ssim-0.834.png}}
     \centering
	 \includegraphics[trim=0.1\imagewidthb{} 0.55\imageheightb{} 0.7\imagewidthb{} 0.25\imageheightb{}, clip, width=0.485\textwidth]{mse_img_0052_mae-105.26_gt-139_pred-33.74_psnr-43.267_ssim-0.834.png}
	 \includegraphics[trim=0.75\imagewidthb{} 0.6\imageheightb{} 0.05\imagewidthb{} 0.2\imageheightb{}, clip, width=0.485\textwidth]{mse_img_0052_mae-105.26_gt-139_pred-33.74_psnr-43.267_ssim-0.834.png}
	\mbox{ Count: 83.2}\\
	\mbox{ PSNR: 43, SSIM: 0.83}
\end{subfigure}
\begin{subfigure}[b]{0.24\textwidth}
\begin{center}
\end{center}
	\includegraphics[width=\textwidth]{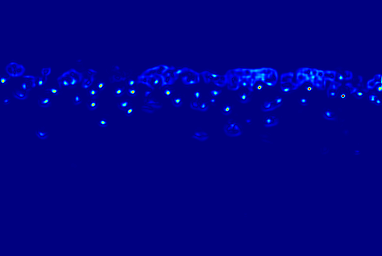}
	\newlength{\imagewidthg}
     \settowidth{\imagewidthg}{\includegraphics{bayesian_img_0052_mae-40.75_gt-139_pred-98.25_psnr-39.131_ssim-0.748.png}}
     \newlength{\imageheightg}
     \settoheight{\imageheightg}{\includegraphics{bayesian_img_0052_mae-40.75_gt-139_pred-98.25_psnr-39.131_ssim-0.748.png}}
     \centering
	 \includegraphics[trim=0.1\imagewidthg{} 0.55\imageheightg{} 0.7\imagewidthg{} 0.25\imageheightg{}, clip, width=0.485\textwidth]{bayesian_img_0052_mae-40.75_gt-139_pred-98.25_psnr-39.131_ssim-0.748.png}
	 \includegraphics[trim=0.75\imagewidthg{} 0.6\imageheightg{} 0.05\imagewidthg{} 0.2\imageheightg{}, clip, width=0.485\textwidth]{bayesian_img_0052_mae-40.75_gt-139_pred-98.25_psnr-39.131_ssim-0.748.png}
	\mbox{ Count: 98.3}\\
	\mbox{ PSNR: 39, SSIM: 0.75}
\end{subfigure}
\begin{subfigure}[b]{0.24\textwidth}
\begin{center}
\end{center}
    \includegraphics[width=\textwidth]{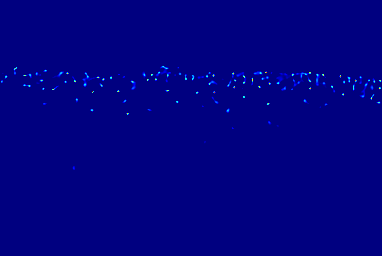}
    \newlength{\imagewidthc}
     \settowidth{\imagewidthc}{\includegraphics{our_img_0052_mae-1.45_gt-139_pred-137.55_psnr-45.891_ssim-0.856.png}}
    \newlength{\imageheightc}
     \settoheight{\imageheightc}{\includegraphics{our_img_0052_mae-1.45_gt-139_pred-137.55_psnr-45.891_ssim-0.856.png}}
     \centering
	 \includegraphics[trim=0.1\imagewidthc{} 0.55\imageheightc{} 0.7\imagewidthc{} 0.25\imageheightc{}, clip, width=0.485\textwidth]{our_img_0052_mae-1.45_gt-139_pred-137.55_psnr-45.891_ssim-0.856.png}
	 \includegraphics[trim=0.75\imagewidthc{} 0.6\imageheightc{} 0.05\imagewidthc{} 0.2\imageheightc{}, clip, width=0.485\textwidth]{our_img_0052_mae-1.45_gt-139_pred-137.55_psnr-45.891_ssim-0.856.png}
    \mbox{Count: 137.6}\\
    \mbox{PSNR: 45, SSIM: 0.86}
    \end{subfigure}
\vskip 0.10in
\begin{subfigure}[b]{0.24\textwidth}
\centering
\includegraphics[width=\textwidth]{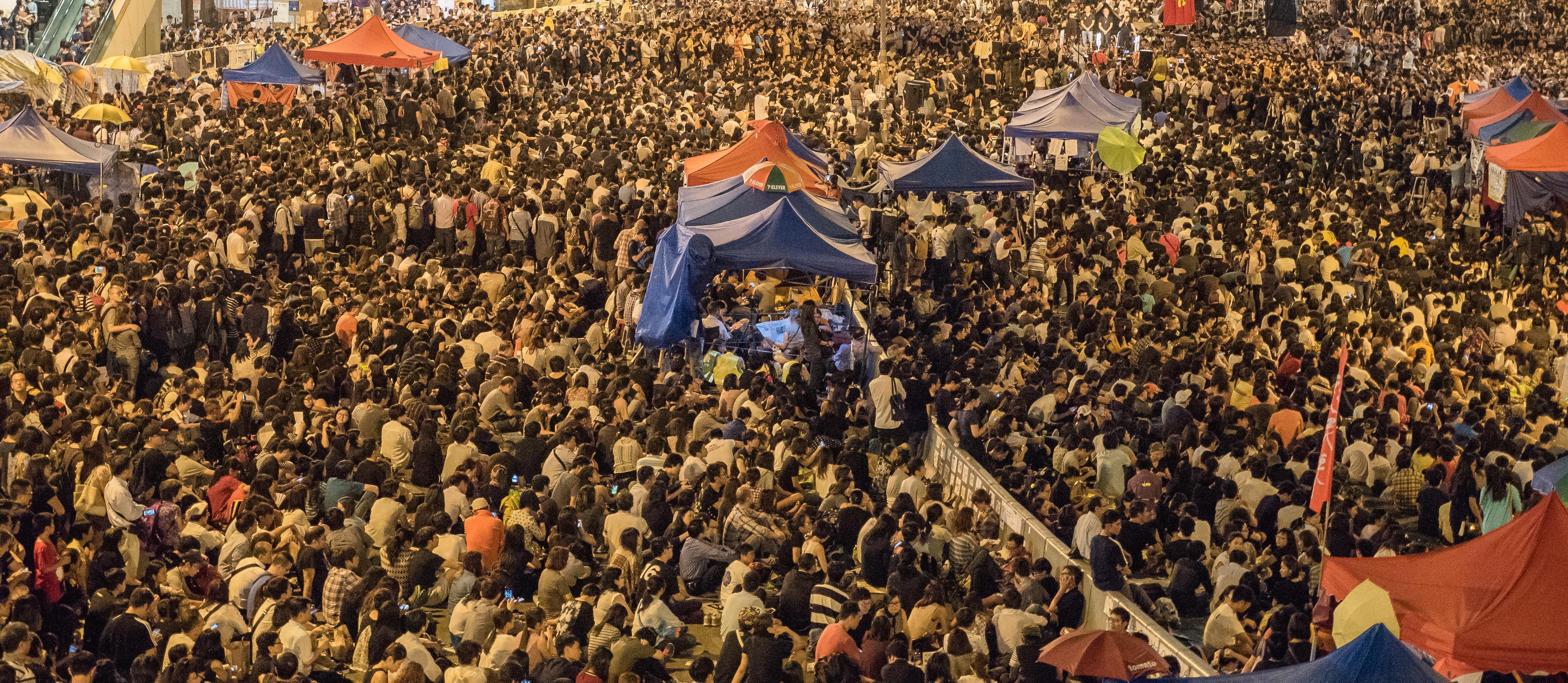}
\newlength{\imagewidthd}
\settowidth{\imagewidthd}{\includegraphics{img_0062.jpg}}
\newlength{\imageheightd}
\settoheight{\imageheightd}{\includegraphics{img_0062.jpg}}
\centering
	 \includegraphics[trim=0.15\imagewidthd{} 0.55\imageheightd{} 0.8\imagewidthd{} 0.35\imageheightd{}, clip, width=0.485\textwidth]{img_0062.jpg}
	 \includegraphics[trim=0.8\imagewidthd{} 0.6\imageheightd{} 0.1\imagewidthd{} 0.2\imageheightd{}, clip, width=0.485\textwidth]{img_0062.jpg}
\mbox{Count: 2638}\\
\mbox{}
\end{subfigure}
\begin{subfigure}[b]{0.24\textwidth}
\centering
	\includegraphics[width=\textwidth]{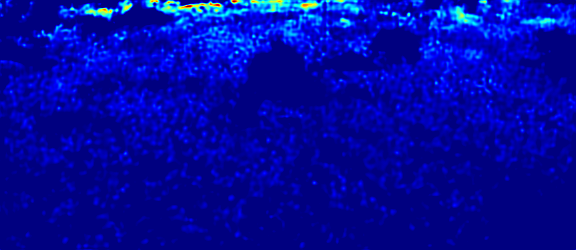}
	\newlength{\imagewidthe}
     \settowidth{\imagewidthe}{\includegraphics{mse_img_0062_mae-882.29_gt-2638_pred-1755.71_psnr-31.306_ssim-0.124.png}}
     \newlength{\imageheighte}
\settoheight{\imageheighte}{\includegraphics{mse_img_0062_mae-882.29_gt-2638_pred-1755.71_psnr-31.306_ssim-0.124.png}}
\centering
	 \includegraphics[trim=0.15\imagewidthe{} 0.55\imageheighte{} 0.8\imagewidthe{} 0.35\imageheighte{}, clip, width=0.485\textwidth]{mse_img_0062_mae-882.29_gt-2638_pred-1755.71_psnr-31.306_ssim-0.124.png}
	 \includegraphics[trim=0.8\imagewidthe{} 0.6\imageheighte{} 0.1\imagewidthe{} 0.2\imageheighte{}, clip, width=0.485\textwidth]{mse_img_0062_mae-882.29_gt-2638_pred-1755.71_psnr-31.306_ssim-0.124.png}
	\mbox{Count: 1755.7}\\
	\mbox{PSNR: 31, SSIM: 0.12}
	\end{subfigure}
\begin{subfigure}[b]{0.24\textwidth}
\centering
	\includegraphics[width=\textwidth]{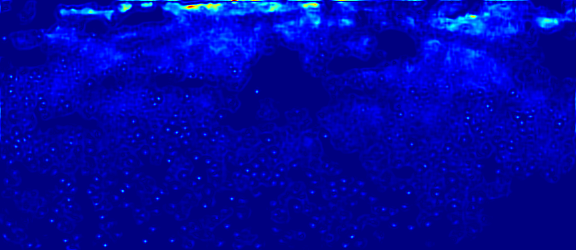}
	\newlength{\imagewidthh}
     \settowidth{\imagewidthh}{\includegraphics{bayesian_img_0062_mae-401.72_gt-2638_pred-2236.28_psnr-30.654_ssim-0.079.png}}
     \newlength{\imageheighth}
\settoheight{\imageheighth}{\includegraphics{bayesian_img_0062_mae-401.72_gt-2638_pred-2236.28_psnr-30.654_ssim-0.079.png}}
\centering
	 \includegraphics[trim=0.15\imagewidthh{} 0.55\imageheighth{} 0.8\imagewidthh{} 0.35\imageheighth{}, clip, width=0.485\textwidth]{bayesian_img_0062_mae-401.72_gt-2638_pred-2236.28_psnr-30.654_ssim-0.079.png}
	 \includegraphics[trim=0.8\imagewidthh{} 0.6\imageheighth{} 0.1\imagewidthh{} 0.2\imageheighth{}, clip, width=0.485\textwidth]{bayesian_img_0062_mae-401.72_gt-2638_pred-2236.28_psnr-30.654_ssim-0.079.png}
	\mbox{Count: 2236.3}\\
	\mbox{PSNR: 30, SSIM: 0.08}
\end{subfigure}
\begin{subfigure}[b]{0.24\textwidth}
\centering
     \includegraphics[width=\textwidth]{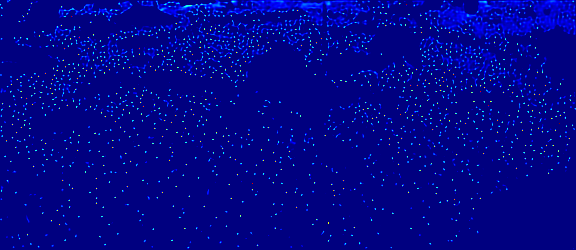}
     \newlength{\imagewidthf}
     \settowidth{\imagewidthf}{\includegraphics{our_img_0062_mae-18.68_gt-2638_pred-2656.68_psnr-37.114_ssim-0.255.png}}
     \newlength{\imageheightf}
\settoheight{\imageheightf}{\includegraphics{our_img_0062_mae-18.68_gt-2638_pred-2656.68_psnr-37.114_ssim-0.255.png}}
\centering
	 \includegraphics[trim=0.15\imagewidthf{} 0.55\imageheightf{} 0.8\imagewidthf{} 0.35\imageheightf{}, clip, width=0.485\textwidth]{our_img_0062_mae-18.68_gt-2638_pred-2656.68_psnr-37.114_ssim-0.255.png}
	 \includegraphics[trim=0.8\imagewidthf{} 0.6\imageheightf{} 0.1\imagewidthf{} 0.2\imageheightf{}, clip, width=0.485\textwidth]{our_img_0062_mae-18.68_gt-2638_pred-2656.68_psnr-37.114_ssim-0.255.png}
     \mbox{Count: 2656.7}\\
     \mbox{PSNR: 37, SSIM: 0.26}
     \end{subfigure}
\caption{{\bf Density map visualization.} Comparison between Pixel-wise loss, Bayesian loss and DM-Count. The pixel-wise  and Bayesian losses fail to localize people well in dense regions. DM-Count is able to localize people both in dense and sparse regions. The Count number, PSNR and SSIM metrics suggest that DM-Count produces more accurate count numbers and better density maps. 
}
\label{fig:visualization}
\vskip 0.10in
\end{figure*}

\begin{figure*}[!t]
\centering
\makebox[0.48\linewidth]{\bf Image}
\makebox[0.48\linewidth]{\bf DM-Count Predicted Result}
\vskip 0.05in
\begin{subfigure}[b]{0.48\textwidth}
\centering
	 \includegraphics[width=\textwidth]{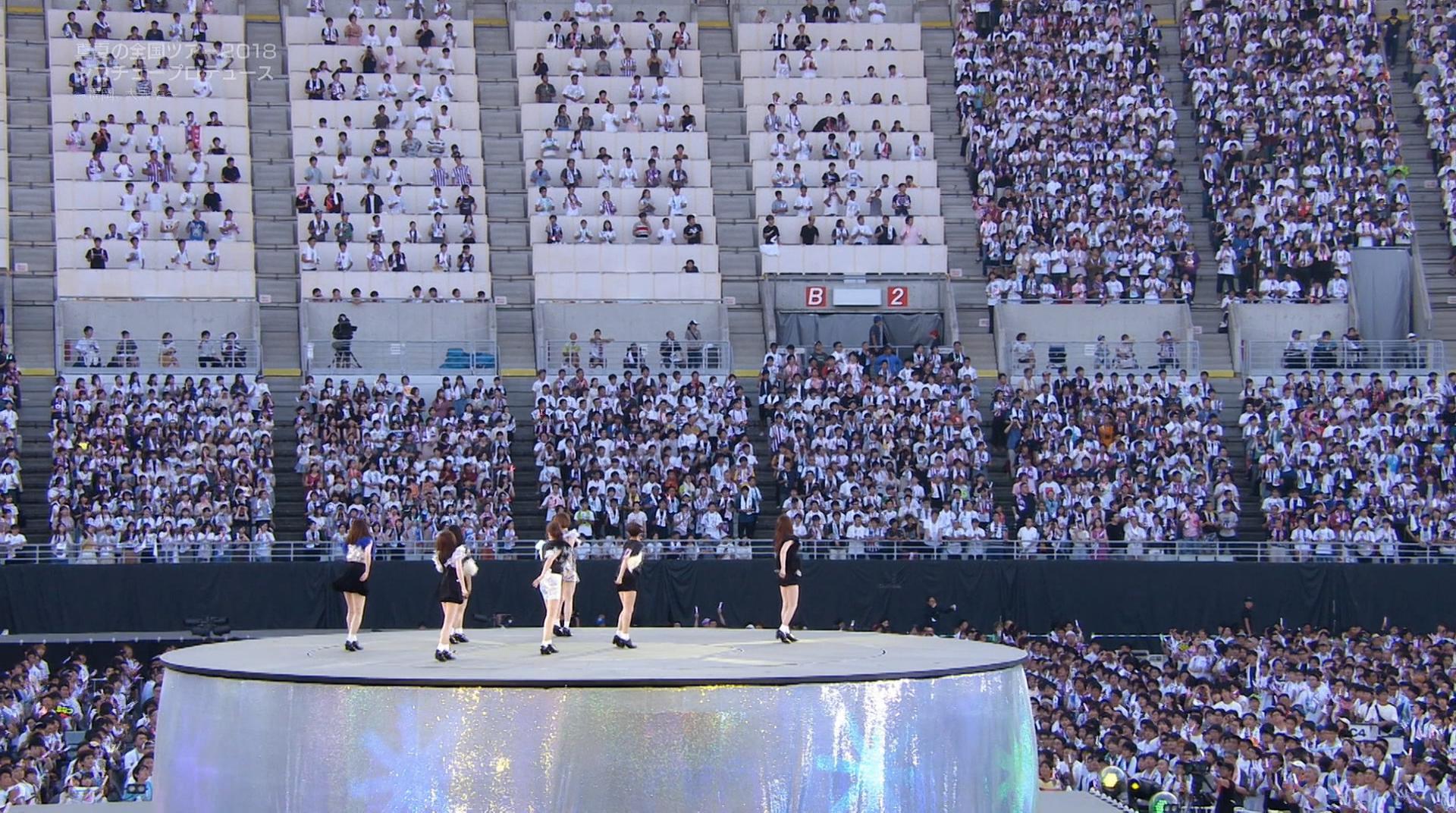}
	 \mbox{Count: 1748}
\end{subfigure}
\begin{subfigure}[b]{0.48\textwidth}
\centering
	 \includegraphics[width=\textwidth]{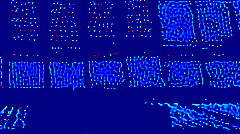}
	 \mbox{Predicted Count: 1634.4} 
\end{subfigure}     
 \vskip 0.10in
 \begin{subfigure}[b]{0.48\textwidth}
 \centering
 	 \includegraphics[width=\textwidth]{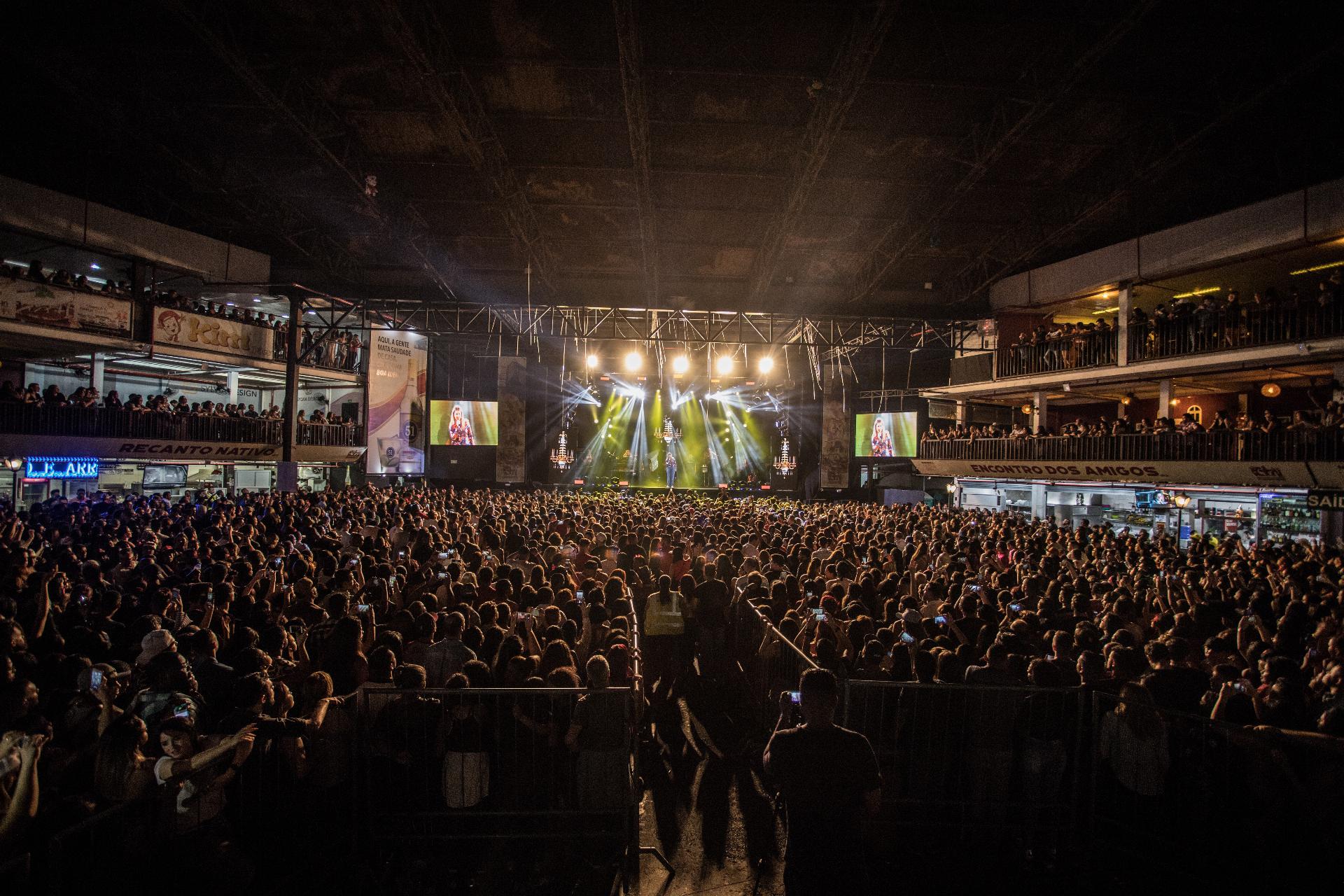}
 	 \mbox{Count: 1270}
 \end{subfigure}
 \begin{subfigure}[b]{0.48\textwidth}
 \centering
 	 \includegraphics[width=\textwidth]{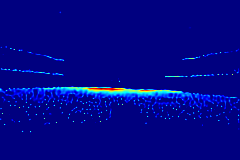}
 	 \mbox{Predicted Count: 1335.3} 
 \end{subfigure}
\caption{{\bf Density map visualization on the NWPU validation set.} 
\label{fig:nwpu_val}}
\end{figure*}

\myheading{Quantitative Results}. Tables~\ref{tab:qnrf_sh_cc} and~\ref{tab:nwpu} compare the performance of DM-Count against various methods. In all experiments, DM-Count outperforms all other methods except CAN under MSE in NWPU (where they are comparable). 
Although we use the same set of hyper-parameters for DM-Count in all experiments, DM-Count still achieves the best performance, suggesting that DM-Count's performance is stable across various datasets. 

DM-Count outperforms the Pixel-wise loss and the Bayesian loss, when used in the same network architecture and training procedure as DM-Count, in all the experiments. This demonstrates the effectiveness of the proposed loss. The pixel-wise loss is much worse than DM-Count in Table \ref{tab:qnrf_sh_cc}. Additionally, even without using a multi-scale architecture as  in ~\cite{cao2018scale, wan2019adaptive}, or a deeper network as  in~\cite{Sam-etal-CVPR17, wang2019learning}, DM-Count still achieves state-of-the-art performance on all four datasets. This indicates the importance of having a good loss function in crowd counting. 

On the large-scale and challenging datasets UCF-QNRF and NWPU, DM-Count significantly outperforms the state-of-the-art methods. Specifically, on the UCF-QNRF dataset, DM-Count reduces the MAE and MSE of the Bayesian loss from 88.7 to 85.6 and from 154.8 to 148.3, respectively. Notably, on the NWPU test set (obtained by submitting to the evaluation server), DM-Count reduces the MAE and NAE by a large margin, from 105.4 to 88.4 in MAE and from 0.203 to 0.169 in NAE. 

\myheading{Qualitative Results}. Fig.~\ref{fig:visualization} shows the predicted density maps of the Pixel-wise loss, the Bayesian loss and DM-Count. This figure demonstrates that: 1) DM-Count produces count numbers that are closer to the ground truth numbers, 2) DM-Count produces much sharper density maps than the Pixel-wise  and Bayesian losses. In Fig.~\ref{fig:visualization}, DM-Count produces much higher PSNRs and SSIMs than the Pixel-wise and Bayesian losses. The average PSNR and SSIM over the whole UCF-QNRF test set for the Pixel-wise loss are 34.79 and 0.43, for the Bayesian loss are 34.55 and 0.42, and for DM-Count are 40.65 and 0.55, respectively. Because the Pixel-wise loss uses the Gaussian smoothed ground truth, it produces blurrier density maps than the real ground truth. This empirically verifies our theoretical analysis of the generalization bound of Gaussian smoothed methods. As shown in the figure, the Pixel-wise and  Bayesian losses are unable to localize people in dense regions. In contrast, DM-Count localizes people well in both dense and sparse regions. Fig. \ref{fig:nwpu_val} shows predicted density maps by DM-Count. The predicted density maps correspond well to crowd densities  in both sparse and dense areas, demonstrating the effectiveness of DM-Count in spatial density estimation.

\begin{table}[t]
    \begin{minipage}{.5\linewidth}
      \centering
         \begin{tabular}{ccc}
        \toprule
        \# Sinkhorn Iters &  MAE    & RMSE\\
        \midrule
        50 & 90.8 & 162.1 \\
        100 &  85.6 & 148.3 \\
        120 &  85.5 & 151.5 \\
        \bottomrule
        \end{tabular}
        \vskip 0.05in
        \caption{{\bf Effect of \# of Sinkhorn iterations.} \label{tab:sinkhorn}}
    \end{minipage}
    \begin{minipage}{.5\linewidth}
      \centering
               \begin{tabular}{ccc}
        \toprule
        Method & MAE          & RMSE\\
        \midrule
        Pixel-wise loss & 144.1 & 232.5 \\
        Bayesian loss  & 108.4 & 187.2  \\
        DM-Count &  \textbf{105.6} & \textbf{181.6} \\
        \bottomrule
        \end{tabular}
        \vskip 0.05in
         \caption{{\bf Robustness to noisy annotations.} \label{tab:noise}}
    \end{minipage} 
\end{table}

\subsection{Ablation Studies}

\myheading{Hyper-parameter study}. We tune $\lambda_1$ and $\lambda_2$ in DM-Count on the UCF-QNRF dataset. First, we fix $\lambda_1$ to 0.1 and tune $\lambda_2$ from 0.01, 0.05 to 0.1. The MAE varies from 85.6, 87.8 to 88.5. As $\lambda_2=0.01$ achieves the best result, we fix $\lambda_2$ to 0.01 and tune $\lambda_1$ from 0.01, 0.05 to 0.1. The MAE varies from 87.2, 86.2 to 85.6. Thus, we set $\lambda_1=0.1$, $\lambda_2=0.01$ and use them on all the datasets.

\myheading{Effect of the number of Sinkhorn iterations}. 
Table~\ref{tab:sinkhorn} lists the results of DM-Count on the UCF-QNRF dataset using different numbers of Sinkhorn iterations. As shown in this table, using a small number of iterations lowers the performance of DM-Count, which indicates that we obtain inaccurate OT solutions. When the number of iterations increases to 100, DM-Count outperforms the previous state-of-the-art. 
The performance plateaued after the number of iterations crossed 100. Therefore, in all of our experiments, we use 100 Sinkhorn iterations for DM-Count.

\begingroup
\setlength{\intextsep}{-5pt}%
\begin{wraptable}{R}{0.42\textwidth}
\centering
\setlength{\tabcolsep}{2pt}
\begin{tabular}{cgdgd}
\toprule
Component & \multicolumn{4}{c}{Combinations}\\
\midrule
Counting loss & \checkmark & \checkmark & \checkmark & \checkmark  \\
OT loss  &  &  &\checkmark & \checkmark \\
TV loss &   &\checkmark & & \checkmark \\
\hline
MAE & 103.1 & 94.9  & 89.3& 85.6 \\
RMSE & 175.9 & 167.4  & 161.3& 148.3 \\
\bottomrule
\end{tabular}
\vskip -0.05in
\caption{{\bf Component analysis} \label{tab:ablation}}
\end{wraptable}

\myheading{Contribution of each component}. The loss in DM-Count is composed of three components, the counting loss, the OT loss and the TV loss. We study the contribution of each component on the UCF-QNRF dataset. Results are listed in Table~\ref{tab:ablation}. As seen in the Table, all components are essential to the final performance. However, the OT loss is the most important component.





\myheading{Robustness to noisy annotations}. Crowd annotation is performed by placing a single dot on a person. Such process is ambiguous and could lead to inevitable annotation errors. We study  how different loss functions perform w.r.t. annotation errors. We add  uniform random noise to the original annotation and train different models with the same noisy annotation. The noise is randomly generated between 0 and 5\% of the image height, and is about 80 pixels on average. As shown in Table~\ref{tab:noise}, the proposed DM-Count is more robust to annotation errors compared to the pixel-wise  Bayesian losses.

\section{Conclusion}
In this paper, we have shown that using the Gaussian kernel to smooth the ground truth dot annotations can hurt the generalization bound of a model when testing on the real ground truth data. Instead, we consider crowd counting as a distribution matching problem and propose DM-Count, based on Optimal Transport, to address this problem. Unlike prior work, DM-Count does not need a Gaussian kernel to smooth the annotated dots. The generalization error bound of DM-Count is tighter than that of the Gaussian smoothed methods. Extensive experiments on four crowd counting benchmarks demonstrated that DM-Count significantly outperforms previous state-of-the-art methods.


\section*{Broader Impact}

Our work is able to more accurately estimate the crowd size in images or videos, such that it can guide crowd control and improve public safety. The estimated crowd count results are interpretable, with better crowd localization, which will increase  transparency of the results for critical applications. In an age when the size of the crowd in various political events often becomes a point of heated dispute, having transparent, accurate and objective counting methods could help the historical record, as well a public acceptance of the estimates.  Our method could potentially be used to protect public health by monitoring social distancing which is becoming increasingly important during the current epidemic. 
This method does not leverage biases in the data.  The proposed method for counting is general, with possible applications to biomedical cell counting, live stock counting and etc. Our work can be adapted to count moving crowds.

\section*{Acknowledgements}

This research was partially supported by US National Science Foundation Award IIS-1763981, the SUNY2020 Infrastructure Transportation Security Center, and Air Force Research Laboratory (AFRL) DARPA FA8750-19-2-1003,  the Partner University Fund, and a gift from Adobe.



{\small
\bibliographystyle{plainnat}
\bibliography{longstrings,pubs,egbib}
}

\end{document}


\def\mA{\mathcal{A}}
\def\mB{\mathcal{B}}
\def\mC{\mathcal{C}}
\def\mD{\mathcal{D}}
\def\mE{\mathcal{E}}
\def\mF{\mathcal{F}}
\def\mG{\mathcal{G}}
\def\mH{\mathcal{H}}
\def\mI{\mathcal{I}}
\def\mJ{\mathcal{J}}
\def\mK{\mathcal{K}}
\def\mL{\mathcal{L}}
\def\mM{\mathcal{M}}
\def\mN{\mathcal{N}}
\def\mO{\mathcal{O}}
\def\mP{\mathcal{P}}
\def\mQ{\mathcal{Q}}
\def\mR{\mathcal{R}}
\def\mS{\mathcal{S}}
\def\mT{\mathcal{T}}
\def\mU{\mathcal{U}}
\def\mV{\mathcal{V}}
\def\mW{\mathcal{W}}
\def\mX{\mathcal{X}}
\def\mY{\mathcal{Y}}
\def\mZ{\mathcal{Z}}

\def\1n{\mathbf{1}_n}
\def\0{\mathbf{0}}
\def\1{\mathbf{1}}

\def\A{{\bf A}}
\def\B{{\bf B}}
\def\C{{\bf C}}
\def\D{{\bf D}}
\def\E{{\bf E}}
\def\F{{\bf F}}
\def\G{{\bf G}}
\def\H{{\bf H}}
\def\I{{\bf I}}
\def\J{{\bf J}}
\def\K{{\bf K}}
\def\L{{\bf L}}
\def\M{{\bf M}}
\def\N{{\bf N}}
\def\O{{\bf O}}
\def\P{{\bf P}}
\def\Q{{\bf Q}}
\def\R{{\bf R}}
\def\S{{\bf S}}
\def\T{{\bf T}}
\def\U{{\bf U}}
\def\V{{\bf V}}
\def\W{{\bf W}}
\def\X{{\bf X}}
\def\Y{{\bf Y}}
\def\Z{{\bf Z}}

\def\a{{\bf a}}
\def\b{{\bf b}}
\def\c{{\bf c}}
\def\d{{\bf d}}
\def\e{{\bf e}}
\def\f{{\bf f}}
\def\g{{\bf g}}
\def\h{{\bf h}}
\def\i{{\bf i}}
\def\j{{\bf j}}
\def\k{{\bf k}}
\def\l{{\bf l}}
\def\m{{\bf m}}
\def\n{{\bf n}}
\def\o{{\bf o}}
\def\p{{\bf p}}
\def\q{{\bf q}}
\def\r{{\bf r}}
\def\s{{\bf s}}
\def\t{{\bf t}}
\def\u{{\bf u}}
\def\v{{\bf v}}
\def\w{{\bf w}}
\def\x{{\bf x}}
\def\y{{\bf y}}
\def\z{{\bf z}}

\def\balpha{\mbox{\boldmath{$\alpha$}}}
\def\bbeta{\mbox{\boldmath{$\beta$}}}
\def\bdelta{\mbox{\boldmath{$\delta$}}}
\def\bgamma{\mbox{\boldmath{$\gamma$}}}
\def\blambda{\mbox{\boldmath{$\lambda$}}}
\def\bsigma{\mbox{\boldmath{$\sigma$}}}
\def\btheta{\mbox{\boldmath{$\theta$}}}
\def\bomega{\mbox{\boldmath{$\omega$}}}
\def\bxi{\mbox{\boldmath{$\xi$}}}
\def\bnu{\mbox{\boldmath{$\nu$}}}                                  
\def\bphi{\mbox{\boldmath{$\phi$}}}
\def\bmu{\mbox{\boldmath{$\mu$}}}

\def\bDelta{\mbox{\boldmath{$\Delta$}}}
\def\bOmega{\mbox{\boldmath{$\Omega$}}}
\def\bPhi{\mbox{\boldmath{$\Phi$}}}
\def\bLambda{\mbox{\boldmath{$\Lambda$}}}
\def\bSigma{\mbox{\boldmath{$\Sigma$}}}
\def\bGamma{\mbox{\boldmath{$\Gamma$}}}
\def\bbE{\mathbb{$E$}}
\def\bbR{\mathbb{$R$}}                                  

\newcommand{\myminimum}[1]{\mathop{\textrm{minimum}}_{#1}}
\newcommand{\mymaximum}[1]{\mathop{\textrm{maximum}}_{#1}}    
\newcommand{\mymin}[1]{\mathop{\textrm{minimize}}_{#1}}
\newcommand{\mymax}[1]{\mathop{\textrm{maximize}}_{#1}}
\newcommand{\mymins}[1]{\mathop{\textrm{min.}}_{#1}}
\newcommand{\mymaxs}[1]{\mathop{\textrm{max.}}_{#1}}  
\newcommand{\myargmin}[1]{\mathop{\textrm{argmin}}_{#1}} 
\newcommand{\myargmax}[1]{\mathop{\textrm{argmax}}_{#1}} 
\newcommand{\myst}{\textrm{s.t. }}

\newcommand{\denselist}{\itemsep -1pt}
\newcommand{\sparselist}{\itemsep 1pt}

\definecolor{pink}{rgb}{0.9,0.5,0.5}
\definecolor{purple}{rgb}{0.5, 0.4, 0.8}   
\definecolor{gray}{rgb}{0.3, 0.3, 0.3}
\definecolor{mygreen}{rgb}{0.2, 0.6, 0.2}

\newcommand{\cyan}[1]{\textcolor{cyan}{#1}}
\newcommand{\red}[1]{\textcolor{red}{#1}}  
\newcommand{\blue}[1]{\textcolor{blue}{#1}}
\newcommand{\magenta}[1]{\textcolor{magenta}{#1}}
\newcommand{\pink}[1]{\textcolor{pink}{#1}}
\newcommand{\green}[1]{\textcolor{green}{#1}} 
\newcommand{\gray}[1]{\textcolor{gray}{#1}}    
\newcommand{\mygreen}[1]{\textcolor{mygreen}{#1}}    
\newcommand{\purple}[1]{\textcolor{purple}{#1}}       

\definecolor{greena}{rgb}{0.4, 0.5, 0.1}
\newcommand{\greena}[1]{\textcolor{greena}{#1}}

\definecolor{bluea}{rgb}{0, 0.4, 0.6}
\newcommand{\bluea}[1]{\textcolor{bluea}{#1}}
\definecolor{reda}{rgb}{0.6, 0.2, 0.1}
\newcommand{\reda}[1]{\textcolor{reda}{#1}}

\def\changemargin#1#2{\list{}{\rightmargin#2\leftmargin#1}\item[]}
\let\endchangemargin=\endlist
                                               
\newcommand{\cm}[1]{}

\newcommand{\mtodo}[1]{{\color{red}$\blacksquare$\textbf{[TODO: #1]}}}
\newcommand{\myheading}[1]{\vspace{1ex}\noindent \textbf{#1}}
\newcommand{\htimesw}[2]{\mbox{$#1$$\times$$#2$}}
\newcommand{\mh}[1]{\textcolor{magenta}{[Minh: {#1}]}}
\newcommand{\ms}[1]{\textcolor{red}{[MS: {#1}]}}

\newif\ifshowsolution
\showsolutiontrue

\ifshowsolution  
\newcommand{\Comment}[1]{\paragraph{\bf $\bigstar $ COMMENT:} {\sf #1} \bigskip}
\newcommand{\Solution}[2]{\paragraph{\bf $\bigstar $ SOLUTION:} {\sf #2} }
\newcommand{\Mistake}[2]{\paragraph{\bf $\blacksquare$ COMMON MISTAKE #1:} {\sf #2} \bigskip}
\else
\newcommand{\Solution}[2]{\vspace{#1}}
\fi

\newcommand{\truefalse}{
\begin{enumerate}
	\item True
	\item False
\end{enumerate}
}

\newcommand{\yesno}{
\begin{enumerate}
	\item Yes
	\item No
\end{enumerate}
}
\newcommand{\Sref}[1]{Sec.~\ref{#1}}
\newcommand{\Eref}[1]{Eq.~(\ref{#1})}
\newcommand{\Fref}[1]{Fig.~\ref{#1}}
\newcommand{\Tref}[1]{Tab.~\ref{#1}}

\definecolor{customgray}{rgb}{0.9, 0.9, 0.9}
\newcolumntype{d}{>{\columncolor[gray]{.7}}c}
\newcolumntype{g}{>{\columncolor[gray]{.9}}c}
\newcolumntype{z}{>{\columncolor[gray]{.9}}l}

\maketitle

\begin{abstract}
  In crowd counting, each person is annotated by a dot and the goal is to count the total number of people in an image. Existing crowd counting methods need to use a Gaussian to smooth each annotated dot or to estimate the likelihood of a pixel given an annotated point. In this paper, we argue that imposing Gaussians to annotations 
  hurts the generalization performance. 
  Instead, we consider crowd counting as a distribution matching problem and propose DIStribution matching for crowd COUNTing (DisCount). In DisCount, we use the Optimal Transport (OT) to measure the similarity between the normalized predicted density map and the normalized ground truth density map. To stabilize OT computation, we include the Total Variation (TV) loss. We present the generalization error bound for DisCount and show that the bound of DisCount is tighter than that of the Gaussian smoothed methods. 
  The Mean Absolute Error (MAE) demonstrates that our method outperforms state-of-the-art methods by a large margin on the world's two largest counting datasets, UCF-QNRF and NWPU, and achieves the state-of-the-art results on the ShanghaiTech and UCF CC 50 datasets. Notably, our method is ranked top 1 on the NWPU benchmark website by reducing approximately 16\% of the MAE of the second best method on the NWPU dataset.
\end{abstract}

\section{Proofs}
\begin{definition}
(Empirical Rademacher complexity \cite{}) Let $\mG$ be a space of functions mapping from $\mI$ to $\mathbb{R}$ and $S = (I_1, ..., I_K) a $ fixed sample of size $K$ with elements
in $\mI$. Then, the empirical Rademacher complexity of $\mG$ with respect to the sample
$S$ is defined as:
\begin{align}
    R_S(\mG) = \mathbb{E}_{\boldsymbol \sigma } \left [ \sup_{g \in \mG} \frac{1}{K} \sum_{k=1}^K \sigma_k g(I_k) \right]
\end{align}
where $\boldsymbol \sigma = (\sigma_1, .., \sigma_K)^{\intercal}$, with $sigma_k$ is independent uniform random variables taking
values in $\{ -1, +1 \}$ the random variables $\sigma_k$ are called Rademacher variables.
\end{definition}

We propose the following lemma

\begin{lemma}
Given two probability measures $\bmu \in \mathbb{R}^n_+$ and $\bnu \in \mathbb{R}^n_{+}$, the optimal transport cost function 
\begin{align}
\mW(\bmu, \bnu) = \min_{\gamma \in \Gamma} ~~& \langle \C, \gamma \rangle, \quad \Gamma = \{ \gamma \in \mathbb{R}^{n \times n}_{+}: \gamma \boldsymbol{1} = \bmu, \gamma^{T} \boldsymbol{1} = \bnu \}
\end{align}
is $\C_{\infty}$ Lipschitz w.r.t. $(\bmu, \bnu)$ under $L_1$ distance, i.e. $\forall (\bmu_1, \bnu_1)$ and $(\bnu_1, \bnu_2)$, we have $|\mW(\bmu_1, \bnu_1) -  \mW(\bmu_2, \bnu_2) | \leq \C_{\infty} \left ( \normone{\bmu_1 - \bmu_2} + \normone{\bnu_1 - 
\bnu_2}\right)$
\end{lemma}

\begin{lemma}
For all $\a \in \mathbb{R}^n_+$, $\normone{\a} \geq 1$ and $\b \in \mathbb{R}^n_+$,  $\normone{\b} \geq 1$, we have 
\begin{align}
    \left \Vert  \frac{\a}{\normone{\a}} - \frac{\b}{\normone{\b}}  \right \Vert_1 \leq 2n \normone{\a - \b}
\end{align}
\end{lemma}

\begin{theorem}
\label{thm: gs_bound}
Assume that $\forall f \in \mathcal{F}$ and $(I, \t) \sim \Tilde{\mathcal{D}}$, we have $\ell(\t, f(I)) \leq B$. Then, for any $0 < \delta < 1$, with probability of at least $1 - \delta$, \\
a) the upper bound of the generalization error is
\begin{align}
    \mR(\mD, f^{\Tilde{S}}_1, \ell_1) \leq \mR(\Tilde{\mD}, f^{\Tilde{\mD}}_1, \ell_1) + 2 n R_K(\mathcal{H}) 
    + 5 B \sqrt{2\log{(8/\delta)}/K} + \mathbb{E}_{(I, \z) \sim \mathcal{D}} \normone{\z - \t}, \nonumber 
\end{align}
b) the lower bound of the generalization error is
\begin{align}
    \mR(\mD, f^{\Tilde{S}}_1, \ell_1) \geq \max \left ( \mathbb{E}_{(I, \z) \sim \mathcal{D}}  \normone{\z - \t} - \mR(\Tilde{\mD}, f^{\Tilde{S}}_1, \ell_1), 0 \right). \nonumber 
\end{align}
\end{theorem}



\section{Introduction}

Image-based crowd counting is an important research problem in computer vision with various applications in many domains including journalism and surveillance. Current state-of-the-art methods~\cite{Xu_2019_ICCV,Cheng_2019_ICCV,Liu_2019_ICCV,Yan_2019_ICCV,Zhao_2019_CVPR,Zhang_2019_CVPR,Jiang_2019_CVPR,Wan_2019_CVPR,Lian_2019_CVPR,Liu_2019_CVPR} treat crowd counting as a densit y map estimation problem, where a deep neural network first produces a 2D crowd density map for a given input image and subsequently estimates the total size of the crowd by summing the density values across all spatial locations of the density map. For images of large crowds, this density map estimation approach has been shown to be more robust than the detection-then-counting approach~\cite{lin2001estimation, li2008estimating, zhao2003bayesian, ge2009marked} because the former is less sensitive to occlusion and it does not need to commit to binarized decisions at an early stage.

For a density map estimation method, a crucial step in its development is the training  of a deep neural network that maps from an input image to the corresponding annotated density map. In all existing crowd counting datasets~\cite{idrees2018composition, zhang2016single, idrees2013multi, wang2020nwpu}, the annotated density map for each training image is a sparse binary mask, where each individual person is marked with a single dot on their head or forehead. The spatial extent of each person is not provided, due to the laborious effort needed for delineating the spatial extent, especially when there is too much occlusion ambiguity. Given training images with dot annotation, training the density map estimation network is equivalent to optimizing the parameters of the network to minimize a differentiable loss function that measures the discrepancy between the predicted density map and the annotated dot-annotation map. Notably, the former is a dense real-value matrix, while the later is a sparse binary matrix. Given the sparsity of the dots, a function that is defined based on the pixel-wise difference between the annotated and predicted density maps is hard to train because the reconstruction loss is heavily imblanced between the 0s and 1s in the sparse binary matrix. One approach to alleviate this problem is to turn each annotated dot into a Gaussian blob such that the ground truth is more balanced and thus the network is easier to train. Almost all prior crowd density map estimation methods~\cite{zhang2019relational, zhang2019attentional, zhang2016single,sam2017switching,li2018csrnet,onoro2016towards,ranjan2018iterative,cao2018scale,wang2019learning,liu2019context,shi2019revisiting,liu2019point,liu2019adcrowdnet} have been developed based on this approach. 
Unfortunately, the performance of the resulting network is highly dependent on the quality of this ``pseudo ground truth'', but it is not trivial to set the right widths for the Gaussian blobs given huge variation in the sizes, scales and shapes of people in a crowded scene.

Recently,~\citet{bayesianCounting} proposed the so-called Bayesian loss to measure the discrepancy between the predicted density map and the annotated density map. This method transforms a binary ground truth annotation map into $C$ ``smoothed ground truth'' maps, where $C$ is count number. Each smoothed ground truth map is the posterior of the probability of an annotation dot given all the pixels in the image. 
Empirically, this method has been shown to outperform other aforementioned approaches. However, there are two major problems with this loss function. First, it also requires a Gaussian  to construct the likelihood function for each annotated dot, which involves setting the kernel width. Second, 
the loss is a underdetermined system and could have infinite number of predicted density maps that are not similar to the ground truth but can make the final loss to be 0. As a consequence, the resulting predicted density map can be very different from the ground truth density map. 


This paper proposes a novel method to address the crowd counting problem. We have the following contributions.

\begin{itemize} [leftmargin=3.5mm,noitemsep,topsep=0pt,parsep=0pt,partopsep=0pt]
\item
We argue that imposing Gaussians to annotations in crowd counting could hurt the generalization performance. Specifically, we theoretically show that the better one model achieves on Gaussian smoothed ground truth, the poorer it generalizes on the real ground truth data.
\item
We consider crowd counting as a distribution matching problem, and propose the \underline{Dis}tribution matching for crowd \underline{Count}ing (DisCount). Unlike previous works, DisCount does not need any Gaussian kernel to smooth ground truth annotations. In DisCount, Optimal Transport (OT) is used to measure the similarity between the normalized predicted density map and the normalized ground truth density map. The Total Variation (TV) loss is introduced to stabilize OT computation. 
\item
We present the generalization error bounds for the count loss, the OT loss, the TV loss and the overall loss in our method. All the bounds are tighter than that of the Gaussian smoothed methods.  
\item
The Mean Absolute Error (MAE) demonstrates that DisCount outperforms state-of-the-art methods by a large margin on the world's two largest counting datasets, UCF-QNRF and NWPU, and achieves the state-of-the-art results on the ShanghaiTech and UCF CC 50 datasets. Notably, our method is ranked top 1 on the NWPU benchmark website by reducing approximately 16\% of the MAE of the second best method on the NWPU dataset.
\end{itemize}

\section{Previous Work}

\subsection{Crowd Counting Methods}
Crowd counting methods can be divided into three categories: detection-then-count, direct count regression, and density map estimation. 
Early methods~\cite{lin2001estimation, li2008estimating, zhao2003bayesian, ge2009marked} tackle the problem by detecting people, heads, or upper bodies in the image. However, accurate detection is difficult for dense crowd. Besides, it also requires bounding box annotation, which is a laborious and ambiguous process due to heavy occlusion. Later methods~\cite{chan2009bayesian,chen2012feature, wang2015deep, chen2013cumulative} avoid the detection problem and directly learn to regress the count from a feature vector. But their results are less interpretable and the dot annotation maps are underutilized. 
Most recent works~\cite{zhang2019relational, zhang2019attentional, zhang2016single,sam2017switching,li2018csrnet,onoro2016towards,ranjan2018iterative,idrees2018composition,cao2018scale,bayesianCounting,wang2019learning,liu2019context,shi2019revisiting,liu2019point,liu2019adcrowdnet,Xu_2019_ICCV,Cheng_2019_ICCV,Liu_2019_ICCV,Yan_2019_ICCV,Zhao_2019_CVPR,Zhang_2019_CVPR,Jiang_2019_CVPR,Wan_2019_CVPR,Lian_2019_CVPR,Liu_2019_CVPR, lu2018class, wan2019adaptive} are based on density map estimation, which has been shown to be more robust than detection-then-count and count regression approaches. 

Crowd density map estimation methods usually define the training loss based on the pixel-wise difference between the Gaussian smoothed ground truth density map and predicted density map. Instead of using one kernel width to smooth the dot annotation, \cite{zhang2016single, idrees2013multi, wan2019adaptive} used adaptive kernel width. The kernel width is selected based on the distance to its nearest neighbors in the spatial domain. Specifically, \cite{idrees2018composition} generated multiple smoothed ground truth density maps on different density levels. The final loss is composed of reconstruction errors from multiple density levels. However, adaptive kernel based method assumes the crowd is evenly distributed, in reality crowd distribution is quite irregular. The Bayesian loss method \cite{bayesianCounting} uses a Gaussian to construct a likelihood function for each annotated dot. However, Bayesian loss cannot predict an correct density because the loss is underdetermined. The detailed analysis can be found in Section 4.2.

\subsection{Optimal Transport}


We propose a novel loss function defined based on Optimal Transport. For a better understanding of the proposed method, we briefly review OT formulation of Monge-Kantorovich in this section. 

Optimal Transport refers to the optimal cost to transform one probability distribution to another. Let $\mX = \{ \x_i | \x_i \in \mathbb{R}^d\}_{i=1}^n$ and $\mY = \{ \y_j | \y_j \in \mathbb{R}^d \}_{j=1}^n$ be two sets of points on the $d$-dimensional vector space. Let $\bmu$ and $\bnu$ be two probability measures defined on $\mX$ and $\mY$, respectively; $\bmu, \bnu \in \mathbb{R}^n_{+}$ and $\1_n^T\bmu = \1_n^T\bnu = 1$ ($\1_n$ is a $n$-dimensional vector of all ones). Let $c: \mX \times \mY \mapsto \mathbb{R}_{+}$ be the cost function for moving from a point in $\mX$ to a point in $\mY$, and $\C$ be the corresponding $n{\times}n$  cost matrix for the two sets of points: $C_{ij} = c(\x_i, \y_j)$. Let $\Gamma$ be the set of all possible ways to transport probability mass from $\mX$ to $\mY$: $\Gamma = \{ \gamma \in \mathbb{R}^{n \times n}_{+}: \gamma \boldsymbol{1} = \bmu, \gamma^{T} \boldsymbol{1} = \bnu \}$. The Monge-Kantorovich's Optimal Transport (OT) distance between $\bmu$ and $\bnu$ is defined as: 
\begin{align}
\mW(\bmu, \bnu) = \min_{\gamma \in \Gamma} ~~& \langle \C, \gamma \rangle. 
\end{align}
Intuitively, if the probability distribution $\bmu$ is viewed as a unit amount of ``dirt'' piled on $\mX$ and $\bnu$ a unit amount of dirt piled on $\mY$, the OT distance is the minimum ``cost'' of turning one pile into the other. The OT distance is a principal metric to quantify the dissimilarity between two probability distributions, also taking into account the distance between ``dirt'' locations. 

The OT distance can also be computed via the dual formulation: 
    \begin{align}
    \label{eq: ot_dual}
\mW(\bmu, \bnu) = \max_{\balpha, \bbeta \in \mathbb{R}^{n}} & \langle \balpha, \bmu \rangle + \langle \bbeta, \bnu \rangle, \quad  \myst  \alpha_i + \beta_j \leq c(\x_i, \y_j),~\forall i, j.
    \end{align}

\section{DisCount: Distribution Matching for Crowd Counting}



We consider crowd counting as a distribution matching problem. In this section, we propose DisCount: Distribution matching for crowd counting. 
A network for crowd counting inputs an image and outputs a map of density values. The final count estimate can be obtained by summing over the predicted density map. DisCount is agnostic to different network architectures. In our experiments, we use the same network as in Bayesian loss~\cite{bayesianCounting}, with implementation details provided in Section 5.2. Unlike all previous density map estimation methods which need to use Gaussians to preprocess ground truth annotations, DisCount does not need any Gaussian to preprocess ground truth annotations. 



Let $\z \in \mathbb{R}^n_{+}$ denote the vectorized binary map for dot-annotation and $\hat{\z} \in \mathbb{R}^n_{+}$ the vectorized predicted density map returned by a neural network. By viewing $\z$ and $\hat{\z}$ as unnormalized density functions, we define the loss function in DisCount with three terms: the count loss, the OT loss, and the Total Variation (TV) loss. The first term measures the difference between the total masses, while the last two measures the difference in the distribution of the normalized density functions. 

\subsection{The Count Loss}

Let $\normone{\cdot}$ denote the $L_1$ norm of a vector, and so $\normone{\z}$, $\normone{\hat{\z}}$ are the ground truth and predicted counts respectively. The goal of crowd counting is to make $\normone{\hat{\z}}$ as close as possible to $\normone{\z}$, and the count loss is defined as the absolute difference between the them: 
\begin{align}
\label{eq: counting_loss}
    \ell_C \left  (\z, \hat{\z} \right) = | \normone{\z} - \normone{\hat{\z}} |. 
\end{align}

\subsection{The Optimal Transport Loss}


Both $\z$ and $\hat{\z}$ are un-normalized density functions, but we can turn them into probability density functions (pdfs) by dividing them by the their respective total mass. Apart from OT, the Kullback-Leibler divergence and Jensen-Shannon divergence can also measure the similarity between two pdfs. However, these measurements do not provide valid gradients to train a network if the source distribution does not overlap with the target distribution \cite{martin2017wasserstein}. Therefore, we propose to use OT in this work. We define the OT loss as follows: 
\begin{align}
\ell_{OT} \left(\z, \hat{\z} \right ) = \mW\left( \frac{\z}{\normone{\z}}, \frac{\hat{\z}}{\normone{\hat{\z}}}\right ) = \left\langle \balpha^*, \frac{\z}{\normone{\z}}  \right\rangle + \left\langle \bbeta^*, \frac{\hat{\z}}{\normone{\hat{\z}}} \right\rangle, \label{eqn:OT4z}
\end{align}
where $\balpha^*$ and $\bbeta^*$ are the solutions of Problem (\ref{eq: ot_dual}). We use the quadratic transport cost, i.e., $c \left (\z(i), \hat{\z}(j) \right) = \norm{\z(i) - \hat{\z}(j)}_2^2$, where $\z(i)$ and $\hat{\z}(j)$ are 2D coordinates of locations $i$ and $j$, respectively. To avoid the division-by-zero error, we add a machine precision to the denominator. 

Since the entries in $\hat{\z}$ are non-negative, the gradient of Eq.~(\ref{eqn:OT4z}) with respect to  $\hat{\z}$ is:
\begin{align}
\label{eq: ot_grad}
	\frac{\partial \ell_{OT}\left(\z, \hat{\z} \right )  }{\partial \hat{\z}} = \frac{\bbeta^*}{\normone{\hat{\z}}} - \frac{ \langle \bbeta^*, \hat{\z} \rangle }{\normone{\hat{\z}}^2}.
\end{align}
This gradient vector can be back-propagated to learn the parameters of the density map estimation network.


\subsection{Total Variation Loss}

In each training iteration, we use the Sinkhorn algorithm \cite{peyre2019computational} to approximate $\balpha^*$ and $\bbeta^*$. The time complexity is $O(n^2 \log n  / \epsilon^2)$ \cite{pmlr-v80-dvurechensky18a}, where $\epsilon$ is the desired optimality gap, i.e., the upper bound for the difference between the returned objective and the optimal objective. Using the Sinkhorn algorithm for optimization, the objective decreases dramatically at the beginning but it converges slowly to the optimal objective in later iterations. In practice, we need to set the maximum number of iterations, and the Sinkhorn algorithm only returns an approximate solution. As a result, using the OT loss with the Sinkhorn algorithm, the predicted density map will be driven to be close to the ground truth density map, but not necessarily the same. The OT loss will approximate well the dense areas of the crowd, but the approximation might be poorer for low density areas of the crowd. To address this issue, we propose to additionally use the Total Variation (TV) loss, which is defined as: 
\begin{align}
\label{eq: tvloss}
\ell_{TV}(\z, \hat{\z}) = \left\Vert \frac{\z}{\normone{\z}} - \frac{\hat{\z}}{\normone{\hat{\z}}}\right\Vert_{TV} = \frac{1}{2} \left\Vert \frac{\z}{\normone{\z}} - \frac{\hat{\z}}{\normone{\hat{\z}}}\right\Vert_1,
\end{align}
The TV loss will also increase the stability of the training procedure. Optimizing the OT loss with the Sinkhorn algorithm is a min-max saddle point optimization procedure, which is similar to the optimization of GAN~\cite{goodfellow2014generative}. The stability of GAN training can be increased by adding a reconstruction loss, as shown in Pix2Pix GAN~\cite{isola2017image}. To this end, the TV loss is similar to the reconstruction loss, and it will also increase the stability of the training procedure.

The gradient of the TV loss with respect to the predicted density map $\hat{\z}$ is: 
\begin{align}
\label{eq: tv_grad}
	\frac{\partial \ell_{TV}\left(\z, \hat{\z} \right )  }{\partial \hat{\z}} = - \frac{1}{2}\left (  \frac{\mbox{sign}(\v)}{\normone{\hat{\z}}} - \frac{ \langle \mbox{sign} (\v), \hat{\z} \rangle }{\normone{\hat{\z}}^2} \right),
\end{align}
where $\v = \z / \normone{\z} - \hat{\z} / \normone{\hat{\z}}$, and $\mbox{sign}(\cdot)$ is the Sign function on each element of a vector.

\subsection{The Overall Objective}

The overall loss function is the combination of the count loss, the OT loss, and the TV loss:
\begin{align}
\ell(\z, \hat{\z}) =  \ell_{C} (\z, \hat{\z}) + \lambda_1 \ell_{OT}( \z, \hat{\z}) + \lambda_2  \normone{\z} \ell_{TV}(\z, \hat{\z}) ,
\end{align}
where $\lambda_1$ and $\lambda_2$ are tunable hyper parameters for the OT and TV losses. To ensure the TV losses have the scale with the count loss, we multiply it with the total count.

 
Given $K$ training images $\{I_k\}_{k=1}^K$ with corresponding dot annotation maps $\{\z_k\}_{k=1}^K$, we will learn a deep neural network $f$ for density map estimation by minimizing:
$L(f) = \frac{1}{K}\sum_{k=1}^{K} \ell(\z_k, f(I_k))$.

\section{Generalization Bounds and Theoretical Analysis}

Let $\mI$ denote the set of images and $\mZ$ the set of dot annotation maps. Let $\mD = \{(I, \z)\} $ be the joint distribution of crowd images and corresponding dot annotated maps. Let $\mH$ be a hypothesis space of functions. Each $h \in \mH$ maps from $I \in \mI$ to each dimension of $\z \in \mZ$. Let $R_K(\mathcal{H})$ denote the Rademacher complexity~\cite{bartlett2002rademacher} for $\mathcal{H}$ with sample size $K$. Let $\mathcal{F} = \mathcal{H} \times \dots \times \mathcal{H}$ ($n$ times) be the space of mappings. Each $f \in  \mF$ maps $I \in \mI$ to $\z \in \mZ$. Let $\t$ be the Gaussian smoothed density map of each $\z \in \mD$, and let $\Tilde{\mD} = \{ (I, \t) \}$ be the joint distribution of $(I, \t)$. Given $f \in \mF$ and a loss function $\ell$, let $\mR(\ell, f) = \mathbb{E}_{(I, \z) \sim \mathcal{D}} [ \ell(\z, f(I))]$ denote the expected risk, and $\hat{\mR}(\ell, f) = \frac{1}{K} \sum_{k=1}^K [ \ell(\z_k, f(I_k))]$ denote the empirical risk on the finite training set $\mD_{K} = \{(I_k, \z_k)\}_{k=1}^{K}$. Accordingly, let $\mR'(\ell, f) = \mathbb{E}_{(I, \t) \sim \Tilde{\mD}}[ \ell(\t, f(I))]$ and $\hat{\mR}'(\ell, f) = \frac{1}{K} \sum_{k=1}^K [ \ell(\t_k, f(I_k))]$. Let $\ell_1(\z, \hat{\z}) = \normone{\z - \hat{\z}}$. Let $\hat{f}$ be the empirical risk minimizer over the training data and $f^*$ be the optimal minimizer over the entire distribution $\mD$: $\hat{f}_{\Delta} = \myargmin{f \in \mF} \mR(\ell_{\Delta}, f) $ and  $f^*_{\Delta} = \myargmin{f \in \mF} \hat{\mR} (\ell_{\Delta}, f)$, where $\Delta \in \{1, C, OT, TV, \emptyset \}$. Accordingly, let $g^*_{\Delta} = \myargmin{f \in \mF} \mR'(\ell_{\Delta}, f)$ and $\hat{g}_{\Delta} = \myargmin{f \in \mF} \hat{\mR}'(\ell_{\Delta}, f) $.

\subsection{Generalization Error Bounds of Gaussian Smoothed Methods}
Gaussian smoothed methods use Gaussiasn smoothed annotations to train networks. Below we give the generalization error bounds of these methods ~\cite{ zhang2016single,li2018csrnet,ranjan2018iterative} using the $\ell_1$ reconstruction loss.
\begin{theorem}
\label{thm: gs_bound}
Assume that $\forall f \in \mathcal{F}$ and $(I, \t) \sim \Tilde{\mathcal{D}}$, we have $\ell(\t, f(I)) \leq B$. Then, for any $0 < \delta < 1$, with probability of at least $1 - \delta$, \\
a) the upper bound of the generalization error is
\begin{align}
    \mR(\ell_1, \hat{g}_1) \leq \mR'(\ell_1, g^*_1) + 2 n R_K(\mathcal{H}) 
    + 5 B \sqrt{2\log{(8/\delta)}/K} + \mathbb{E}_{(I, \z) \sim \mathcal{D}} \normone{\z - \t} \nonumber 
\end{align}
b) the lower bound of the generalization error is
\begin{align}
    \mR(\ell_1, \hat{g}_1) \geq \max \left ( \mathbb{E}_{(I, \z) \sim \mathcal{D}}  \normone{\z - \t} - \mR'(\ell_1, g^*_1) - 2 n R_K(\mathcal{H}) 
    - 5 B \sqrt{2\log{(8/\delta)}/K}, 0 \right) \nonumber 
\end{align}
\end{theorem}


In the above theorem, as the number of samples $K$ grows to infinity, $2 n R_K(\mathcal{H})$ and $5 B \sqrt{2\log{(8/\delta)}/K}$ decrease to 0. Theorem \ref{thm: gs_bound} a) shows the upper bound (worst case) of the expected risk on real ground truth data using empirical minimizer trained on pseudo ground truth data $\mR(\ell_1, \hat{g}_1)$ will not exceed $\mR'(\ell_1, g^*_1) + \mathbb{E}_{(I, \z) \sim \mathcal{D}} \normone{\z - \t}$ given sufficient training data. Theorem \ref{thm: gs_bound} b) shows the lower bound (best case) of $\mR(\ell_1, \hat{g}_1)$ is not smaller than $\max ( \mathbb{E}_{(I, \z) \sim \mathcal{D}}  \normone{\z - \t} - \mR'(\ell_1, g^*_1), 0) $ given sufficient training data. This means that the better performance we achieve using the pseudo ground truth data (the smaller $\mR'(\ell_1, g^*_1$)), the larger expected risk $\mR(\ell_1, \hat{g}_1)$. $\mR(\ell_1, \hat{g}_1)$ can be as large as $\mathbb{E}_{(I, \z) \sim \mathcal{D}} \normone{\z - \t}$ when $\mR'(\ell_1, g^*_1) = 0$. This is undesirable because we want the risk $\mR(\ell_1, \hat{g}_1)$ evaluated using the real ground truth to be 0 as well. 



\subsection{The Underdetermined Bayesian Loss}
The Bayesian Loss~\cite{bayesianCounting} is 
\begin{align}
    \ell_{Bayesian}(\z, \hat{\z}) = \sum_{i=1}^{C_{\z}} \lvert 1 - \langle \p_i, \hat{\z} \rangle \rvert, \quad \p_i = \frac{\mN (\q_i, \sigma^2 \1_{2\times2})}{\sum_{i=1}^{C_{\z}} \mN (\q_i, \sigma^2 \1_{2\times2})}
\end{align}
where $C_{\z}$ is number of count of $\z$, and $\mN (\q_i, \sigma^2 \1_{2\times2})$ is a Gaussian distribution centered at $\q_i$ with variance $\sigma^2\1_{2\times2}$. $\q_i$ is the $i^{th}$ annotated dot in $\z$. The dimension of $\p_i$ or $\z$ is the dimension of the density map, $n$. However, since the number of annotated dots $C_{\z}$ is less than $n$, the Bayesian loss is Underdetermined. For a ground truth annotation $\z$, there are infinite many $\hat{\z}$ that can lead to $\ell_{Bayesian}(\z, \hat{\z}) = 0$ and $\hat{\z} \ne \z$. Therefore, Bayesian loss will not be able to learn the correct annotation map. 


\subsection{The Generalization Error Bounds of the Losses in the Proposed Method}
We give the generalization error bounds of the losses in the proposed method in the following theorem. 
\begin{theorem}
\label{thm: proposed_bounds}
Assume that $\forall f \in \mathcal{F}$ and $(I, \z) \sim \mathcal{D}$, we have $\normone{\z} > 1$, $\normone{f(I)} > 1$ (can be satisfied by adding a dummy dimension with value of 1  to both $\z$ and $f(I)$) and $\ell_C(\z, f(I)) \leq B$. Then, for any $0 < \delta < 1$, with probability of at least $1 - \delta$\\
a) the generalization error bound of the count loss is
\begin{align}
    \mR(\ell_C, \hat{f}_C) \leq \mR(\ell_C, f^*_C)  + 2 n R_K(\mathcal{H}) + 5 B \sqrt{2\log{(8/\delta)}/K} \nonumber
\end{align}
b) the generalization error bound of the OT loss is 
\begin{align}
    \mR(\ell_{OT}, \hat{f}_{OT}) \leq \mR(\ell_{OT}, f^*_{OT}) + 4 \C_{\infty} n^2 R_K(\mathcal{H}) + 5\C_{\infty}\sqrt{2\log{(8/\delta)}/K} \nonumber
\end{align}
c) the generalization error bound of the TV loss is 
\begin{align}
    \mR(\ell_{TV}, \hat{f}_{TV}) \leq \mR(\ell_{TV}, f^*_{TV}) + n^2 R_K(\mathcal{H}) + 5 \sqrt{2\log{(8/\delta)}K} \nonumber
\end{align}
d) the generalization error bound of the overall loss is 
\begin{align}
\nonumber
    \mR(\ell, \hat{f}) \leq \mR(\ell, f^*) + (2n + 4 \lambda_1 \C_{\infty} n^2 + \lambda_2 C n^2) R_K(\mathcal{H})
    + 5(B + 4\lambda_1 \C_{\infty} + \lambda_2 C )\sqrt{2\log{(8/\delta)}K} \nonumber
\end{align}
where $\C_{\infty}$ is the maximum cost in the cost matrix in OT, and $C = \sup \{ \normone{\z} ~|~ \forall (I, \z) \sim \mathcal{D} \}$ is the maximum counting number over a dataset.
\end{theorem}

In the above theorem, as $K$ grows, $R_K(\mH)$ and $\sqrt{2\log{(1/\delta)}K}$ decreases. All the expected risks using the empirical minimizers $\mR(\ell_{\Delta}, \hat{f}_{\Delta})$ converges to the expected risks using optimal minimizers $\mR(\ell_{\Delta}, f^*_{\Delta})$, $\Delta \in \{ C, OT, TV, \emptyset \}$, respectively. This means that all the upper bounds are tight. In addition, all the upper bounds are tighter than the upper bound of the Gaussian smoothed methods shown in Theorem \ref{thm: gs_bound} a). The bound of the OT loss in Theorem \ref{thm: proposed_bounds} b) is related to the maximum transport cost $\C_{\infty}$. Therefore, we need to use smaller transport cost in OT for better generalization performance. The coefficient of $R_K(\mH)$ for the count loss is $O(n)$, and for the OT loss and the TV loss is $O(n^2)$. This means that for lager image size, we need more number of images to train. The number is linear in size of $\z$ using solely the count loss, and quadratic using solely the OT loss or the TV loss. When using all the three losses, we need to set $\lambda_1$ and $\lambda_2$ smaller in order to balance the three losses.

\section{Experiments}

\begin{figure*}[!t]
\begin{subfigure}[b]{0.24\textwidth}
\centering
{\bf Target Density Map}
	 \includegraphics[width=\textwidth]{images/toy/gt_keypoints_downsample.pdf}
	 \mbox{134}
\end{subfigure}
\begin{subfigure}[b]{0.24\textwidth}
\centering
{\bf Pixel-wise Loss}
	 \includegraphics[width=\textwidth]{images/toy/mse_mae-1.0273_gt-134.0_pred-132.9727.png}
	 \mbox{132.97}
\end{subfigure}
\begin{subfigure}[b]{0.24\textwidth}
\centering
{\bf Bayesian Loss}
	 \includegraphics[width=\textwidth]{images/toy/bayesian_mae-0.6099_gt-134.0_pred-133.3901.png}
	 \mbox{133.39}
\end{subfigure}
\begin{subfigure}[b]{0.24\textwidth}
\centering
{\bf DisCount (ours)}
	 \includegraphics[width=\textwidth]{images/toy/ot_mae-0.0136_gt-134.0_pred-133.9864.png}
	 \mbox{133.98}
\end{subfigure}     
     
\caption{{\bf Comparison of different methods on toy data.} The pixel-wise loss generates blurry density map with higher count error. The predicted density map using Bayesian loss is quite different from the ground truth: it has higher values where this is no annotation. Our proposed loss is able to produce accurate results both on crowd estimation and localization. \mh{Can you use the same color scheme for all methods including the GT? Hmm, red on blue is not good. Can you find the appropriate color scheme for this? } 
\label{fig:toy}}
\end{figure*}

\subsection{Performance on Toy Data}

To understand the empirical behavior of different methods, we consider a toy problem where the task is to move a source density map $\hat{\z}$ to a target density map $\z$ using different losses. The source density map is updated iteratively using gradient descent: $\hat{\z} \leftarrow \hat{\z} - \eta \frac{\partial \ell(\z, \hat{\z})}{\partial \hat{\z}}$. We consider three different methods: pixel-wise loss method, Bayesian loss method~\cite{bayesianCounting}, and the DisCount. The Gaussian smoothed target density map is denoted as $\t$ and pixel-wise loss is the Mean Squared Error (MSE) between $\z$ and $\hat{\z}$. We set $\eta=10^{-5}$ and update $\hat{\z}$ until convergence. The target density map $\z$ is generated by randomly cropping a 64$\times$64 patch from existing annotation map. The source density map $\hat{\z}$ is initialized from a uniform distribution between 0 to 0.01. All three methods start from the same source density map. Fig.~\ref{fig:toy} visualizes the final values of $\hat{\z}$ at convergence. The pixel-wise loss yields a blurry density map with a higher count error than DisCount. The Bayesian loss performs better than the pixel-wise loss in terms of count error, but the resulting density map is quite different from the target, with high values at many locations where no dots are annotated. This confirms our analysis that the Bayesian loss corresponds to an underdetermined system such that the output density map could be very different from the target density map. In contrast, the proposed DisCount is able to produce more accurate crowd count and more similar density as the ground truth annotation than the other two methods.

\mh{Is this the same for real data as well? How do the density maps look for training data with different losses?}

\subsection{Performance on Benchmark Datasets}

\myheading{Datasets}. We perform experiments on four challenging crowd counting datasets: UCF-QNRF, NWPU, ShanghaiTech, and UCF-CC-50. The \textbf{UCF-QNRF} dataset~\cite{idrees2018composition}  contains 1201 train and 334 test images of variable sizes, with 1.25 million dot annotations. The number of people in each image varies from 49 to 12,865 with the average being 815. The \textbf{ShanghaiTech}~\cite{zhang2016single} dataset consists of two parts. Part A contains 482 images collected from the web, and Part B contains 716 images collected on the streets of Shanghai. Images in Part A contains more people than images in Part B; the average counts for Part A and Part B are 501 and 124, respectively. Part A has 300 train and 182 test images, while Part B has 400 train and 316 test images. The \textbf{UCF-CC-50} dataset~\cite{idrees2013multi} contains 50 images from the web, with the minimum, average, and maximum numbers of people being 94, 1280, and 4532. Given the small size of the dataset, we  perform five-fold cross validation and report the average result, as also done in previous works~\cite{zhang2015cross, zhang2016single, Sam-etal-CVPR17, ranjan2018iterative, shi2019revisiting}. The \textbf{NWPU}~\cite{wang2020nwpu} dataset is the latest and largest crowd counting dataset comprising of 5,109 images taken from the web and video sequences. The dataset consists of more than $2.1$ million annotated people, and the count varies between 0 and 20,033. This dataset covers a wider range of lighting conditions, image counts, and crowd appearance. The dataset is split into 3,109 training, 500 validation, and 1500 test images. The ground truth counts for test images are not released, and the results on the test set must be obtained by submitting to the evaluation server \url{https://www.crowdbenchmark.com/nwpucrowd.html}. We report performance on both the validation and test sets.

 \myheading{Evaluation metrics}. Following previous works, we use the following metrics: Mean Absolute Error (MAE), Root Mean Squared Error (RMSE), and mean Normalized Absolute Error (NAE). Given $K$ test images, and let $C_k$ and $\hat{C}_k$ be the ground truth and predicted counts for the $k^{th}$ image. 
 $MAE = \frac{1}{K}\sum_{k=1}^{K}  \lvert C_k - \hat{C}_k  \rvert;
RMSE = \sqrt[]{\frac{1}{K}\sum_{k=1}^{K} (C_k - \hat{C}_k)^2};
NAE = \frac{1}{K}\sum_{k=1}^{K} \frac{ \lvert C_k - \hat{C}_k \rvert}{C_k}$.

\myheading{Implementation details}. For a fair comparison, we use the same network as in Bayesian loss~\cite{bayesianCounting}. The network consists of standard feature extraction backbone, followed by a density map estimation head. VGG-19 is used as the backbone, with the removal of the last pooling layer and the subsequent fully connected layers. The output of the backbone is upsampled to 1/8 of the input image size by bilinear interpolation. The density map prediction head has two $3{\times}3$ convolutional layers with 256 and 128 channels, followed by one $1{\times}1$ convolutional layer.

Due to large sizes of images in UCF-QNRF and NWPU, following~\cite{bayesianCounting, wang2020nwpu, idrees2018composition} we limit the shorter size of image within 2048 and 1920 for UCF-QNRF and NWPU respectively. Random crops are taken for training. We use the same crop sizes as previous works~\cite{bayesianCounting, wang2020nwpu}: 256 for ShanghaiTech Part A and UCF-CC-50, 512 for ShanghaiTech Part B and UCF-QNRF, and 384 for NWPU.

In all experiments, we set $\lambda_1=0.1, \lambda_2 = 0.01$, and the entropic regularization parameter in Sinkhorn to 10.
We implement the Sinkhorn algorithm in PyTorch~\cite{paszke2019pytorch} to speed up the process. The number of Sinkhorn iterations is set to 100. On average, the OT computation time is 25\textit{ms} for each image.


\subsubsection{Results}

\begin{table}[!tb]
\centering
\begin{tabular}{lcc|cc|cc|cc}
\toprule
         & \multicolumn{2}{c}{UCF-QNRF} & \multicolumn{2}{c}{ShanghaiTech A}  & \multicolumn{2}{c}{ShanghaiTech B}  & \multicolumn{2}{c}{UCF-CC-50} \\
               & MAE          & RMSE & MAE          & RMSE & MAE          & RMSE & MAE          & RMSE  \\
\midrule
Crowd CNN~\cite{zhang2015cross} &-&-  &        181.8      &  277.7           &   32.0           &  49.8       & 467.0 & 498.5    \\
MCNN~\cite{zhang2016single} & 277 & 426 &     110.2     &    173.2         &  26.4            &  41.3 & 377.6 & 509.1  \\
CMTL~\cite{sindagi2017cnn}& 252 & 514 & 101.3 & 152.4 & 20.0 & 31.1 & 322.8 & 341.4  \\
Switch CNN ~\cite{Sam-etal-CVPR17} & 228 & 445  & 90.4           &  135.0           &     21.6         &     33.4   & 318.1& 439.2      \\
CP-CNN~\cite{sindagi2017generating}  &-&- & 73.6 & 106.4 &20.1 & 30.1 & 295.8 & 320.9  \\
IG-CNN~\cite{babu2018divide} &-&- & 72.5 & 118.2 & 13.6 & 21.1  & 291.4 & 349.4 \\
ic-CNN~\cite{ranjan2018iterative}  & - & - & 68.5 &116.2 & 10.7& 16.0  &260.9 & 365.5\\
CSR Net~\cite{li2018csrnet} & -&- & 68.2 & 115.0 & 10.6 & 16.0  &  266.1 & 397.5\\
SANet~\cite{cao2018scale}  & -&-  &  67.0 & 104.5 & 8.4 & 13.6 &  258.4 & 334.9  \\
CL-CNN~\cite{idrees2018composition} & 132 & 191 &-&-&-&-&-&- \\
PACNN ~\cite{shi2019revisiting} & -&-  & 62.4 & 102.0 & 7.6 & 11.8 & 241.7 & 320.7 \\
CAN ~\cite{liu2019context} & 107 & 183 & 62.3 & 100.0 & 7.8 & 12.2  & 212.2 & 243.7   \\
SFCN ~\cite{wang2019learning} & 102  &   171 & 64.8  & 107.5  &7.6   &  13.0  & 214.2 & 318.2\\
ANF~\cite{zhang2019attentional} & 110 & 174 & 63.9 & 99.4 &  8.3 & 13.2 &250.2 &340.0  \\
Wan \textit{et al.}~\cite{wan2019adaptive} & 101 & 176& 64.7 & 97.1 & 8.1 & 13.6 &-&- \\
\midrule
pixel-wise Loss~\cite{bayesianCounting} & 106.8 & 183.7 & 68.6 & 110.1 & 8.5 & 13.9 & 251.6 & 331.3 \\
Bayesian Loss~\cite{bayesianCounting} & 88.7  & 154.8 & 62.8 & 101.8  & 7.7 & 12.7 &229.3 & 308.2 \\
DisCount (ours)  & \textbf{85.6} & \textbf{148.3} & \textbf{59.7} & \textbf{95.7} & \textbf{7.4} & \textbf{11.8}   & \textbf{211.0} & \textbf{291.5} \\
\bottomrule
\end{tabular}
\vskip 0.05in
\caption{{\bf Results on the UCF-QNRF, Shanghai Tech, and UCF-CC-50 datasets}. \label{tab:qnrf_sh_cc}}
\end{table}

\begin{table}[!tb]
\centering
\begin{tabular}{lcccccccc}
\toprule
         & Backbone  & \multicolumn{2}{c}{Validation set} & \multicolumn{3}{c}{Test set} \\
         \cmidrule(lr){3-4} \cmidrule(lr){5-7} 
             &  &  MAE & RMSE &  MAE   & RMSE  & NAE\\
\midrule
MCNN~\cite{zhang2016single} & FS & 218.5 & 700.6 & 232.5 & 714.6 & 1.063  \\
CSR net~\cite{li2018csrnet} & VGG-16 & 104.8 & 433.4 & 121.3 & 387.8 & 0.604\\
PCC-Net-VGG & VGG-16 & 100.7 & 573.1 & 112.3 & 457.0 & 0.251 \\
CAN ~\cite{liu2019context} & VGG-16 & 93.5 & 489.9 & 106.3 & \textbf{386.5} & 0.295 \\
SCAR~\cite{gao2019scar} & VGG-16 & 81.5 & 397.9 & 110.0 & 495.3 & 0.288\\
Bayesian Loss~\cite{bayesianCounting} &  VGG-19 & 93.6 & 470.3 & 105.4 & 454.2 & 0.203 \\
SFCN & ResNet-101 & 95.4 & 608.3 & 105.4 & 424.1 & 0.254 \\
\midrule
DisCount (ours) &  VGG-19  & \textbf{70.5} & \textbf{357.6} & \textbf{88.4} & 388.6 &  \textbf{0.169}  \\
\bottomrule
\end{tabular}
\vskip 0.05in
\caption{{\bf Results of various methods on the NWPU validation and test sets}. \label{tab:nwpu}}
\end{table}

Tables~\ref{tab:qnrf_sh_cc} and~\ref{tab:nwpu} compare the performance of DisCount against various crowd counting methods, including the state-of-the-art method: Bayesian loss. In all the experiments shown in Tables ~\ref{tab:qnrf_sh_cc} and~\ref{tab:nwpu}, DisCount outperforms all the other methods except CAN under MSE (In fact, DisCount is comparable to CAN under MSE). This shows the superiority of DisCount. Also, although we use the same set of hyper-parameters for DisCount in all the experiments, DisCount still achieves the best performance in almost all the experiments. This shows that DisCount's performance is stable across various datasets and hence it is easy to use. 

When comparing DisCount against the Pixel-wise loss and the Bayesian loss, which have same network architecture and training procedure as DisCount, DisCount beats them in all the experiments. This demonstrates the loss in DisCount is better than the Pixel-wise loss and the Bayesian loss. Additionally, without using a multi-scale architecture as used in {\color{blue} [references]}, or a deeper network as used in {\color{blue} [references]}, DisCount still achieves state-of-the-art performance on all the four datasets. This indicates that a good loss function in crowd counting may be more important than the network architecture. 

On the world's two largest and challenging datasets UCF-QNRF and NWPU, DisCount significantly outperforms the state-of-the-art methods. Specifically, on thte UCF-QNRF dataset, DisCount reduces the MAE and MSE of the Bayesian loss from 88.7 to 85.6 and from 154.8 to 148.3, respectively. Notably, on the NWPU test set (obtained by submitting to the evaluation server), DisCount reduces the MAE and NAE by a large margin, from 105.4 to 88.4 in MAE and from 0.203 to 0.169 in NAE. \textbf{Our method is ranked number one on the NWPU leader board.} 

Fig~\ref{fig:visualization} visualizes the predicted density map using our loss. The proposed loss is able to generate sharp and accurate density estimations. 
\mh{What does ``sharp'' mean in this case? Is it in relative or absolute sence? Can you measure the sharpness quantitatively. This might be a good numbmer to report.}

{\color{blue}{will use PSNR, SSIM to evalute the sharpness.} }
To evaluate the sharpness of the output density map, we measure two popular criterias, Peak Signal-to-Noise Ratio (PSNR)~\cite{} and Structural Similarity in Image (SSIM)~\cite{} between predicted density map and ground truth sparse annotation map. higher PSNR and SSIM implies the more similarity between predicted and ground truth map. The PSNR of our method is 40.653, where Bayesian method is 34.554. The SSIM of our method is 0.552, and 0.425 for Bayesian loss. Our method significantly output Bayesian method on these two metrics.



 

\subsection{Ablation Studies}

\begingroup
\setlength{\intextsep}{-3pt}%
\begin{wraptable}{R}{0.43\textwidth}
\centering
\setlength{\tabcolsep}{2pt}
\begin{tabular}{lgdgd}
\toprule
Component & \multicolumn{4}{c}{Combinations}\\
\midrule
Count loss & \checkmark & \checkmark & \checkmark & \checkmark  \\
OT loss  &  &  &\checkmark & \checkmark \\
TV loss &   &\checkmark & & \checkmark \\
\hline
MAE & 103.1 & 94.9  & 89.3& 85.6 \\
RMSE & 175.9 & 167.4  & 161.3& 148.3 \\
\bottomrule
\end{tabular}
\vskip -0.05in
\caption{{\bf Ablation study on UCF-QNRF} \label{tab:ablation}}
\end{wraptable}

The loss in DisCount is composed of three components, the count loss, the OT loss and the TV loss. We study the contribution of each component on the UCF-QNRF dataset. The result is shown in Table~\ref{tab:ablation}. As can be seen from the Table, all components are essential to final performance. Specifically, OT loss is the most important component.

\begin{figure*}[!t]


\begin{subfigure}[b]{0.33\textwidth}
\centering
{\bf Image}
	 \includegraphics[width=\textwidth]{images/img_0052.jpg}
	 \newlength{\imagewidtha}
     \settowidth{\imagewidtha}{\includegraphics{images/img_0052.jpg}}
     \newlength{\imageheighta}
     \settoheight{\imageheighta}{\includegraphics{images/img_0052.jpg}}
	 \includegraphics[trim=0.1\imagewidtha{} 0.55\imageheighta{} 0.7\imagewidtha{} 0.25\imageheighta{}, clip, width=0.49\textwidth]{images/img_0052.jpg}
	 \includegraphics[trim=0.75\imagewidtha{} 0.6\imageheighta{} 0.05\imagewidtha{} 0.2\imageheighta{}, clip, width=0.49\textwidth]{images/img_0052.jpg}
	 
	 \mbox{Count: 139}
\end{subfigure}
\begin{subfigure}[b]{0.33\textwidth}
\centering
{\bf Bayesian}
	\includegraphics[width=\textwidth]{images/bayesian_img_0052_mae-40.75_gt-139_pred-98.25_psnr-39.131_ssim-0.748.png}
	\newlength{\imagewidthb}
     \settowidth{\imagewidthb}{\includegraphics{images/bayesian_img_0052_mae-40.75_gt-139_pred-98.25_psnr-39.131_ssim-0.748.png}}
     \newlength{\imageheightb}
     \settoheight{\imageheightb}{\includegraphics{images/bayesian_img_0052_mae-40.75_gt-139_pred-98.25_psnr-39.131_ssim-0.748.png}}
	 \includegraphics[trim=0.1\imagewidthb{} 0.55\imageheightb{} 0.7\imagewidthb{} 0.25\imageheightb{}, clip, width=0.49\textwidth]{images/bayesian_img_0052_mae-40.75_gt-139_pred-98.25_psnr-39.131_ssim-0.748.png}
	 \includegraphics[trim=0.75\imagewidthb{} 0.6\imageheightb{} 0.05\imagewidthb{} 0.2\imageheightb{}, clip, width=0.49\textwidth]{images/bayesian_img_0052_mae-40.75_gt-139_pred-98.25_psnr-39.131_ssim-0.748.png}
	 
	\mbox{Count: 98.3, PSNR: 39}
\end{subfigure}
\begin{subfigure}[b]{0.33\textwidth}
\centering
{\bf Our DisCount}
    \includegraphics[width=\textwidth]{images/our_img_0052_mae-1.45_gt-139_pred-137.55_psnr-45.891_ssim-0.856.png}
    \newlength{\imagewidthc}
     \settowidth{\imagewidthc}{\includegraphics{images/our_img_0052_mae-1.45_gt-139_pred-137.55_psnr-45.891_ssim-0.856.png}}
    \newlength{\imageheightc}
     \settoheight{\imageheightc}{\includegraphics{images/our_img_0052_mae-1.45_gt-139_pred-137.55_psnr-45.891_ssim-0.856.png}}
	 \includegraphics[trim=0.1\imagewidthc{} 0.55\imageheightc{} 0.7\imagewidthc{} 0.25\imageheightc{}, clip, width=0.49\textwidth]{images/our_img_0052_mae-1.45_gt-139_pred-137.55_psnr-45.891_ssim-0.856.png}
	 \includegraphics[trim=0.75\imagewidthc{} 0.6\imageheightc{} 0.05\imagewidthc{} 0.2\imageheightc{}, clip, width=0.49\textwidth]{images/our_img_0052_mae-1.45_gt-139_pred-137.55_psnr-45.891_ssim-0.856.png}
	
    \mbox{Count: 137.6, PSNR: 45}
    \end{subfigure}

\begin{subfigure}[b]{0.33\textwidth}
\centering
\includegraphics[width=\textwidth]{images/img_0062.jpg}
\newlength{\imagewidthd}
\settowidth{\imagewidthd}{\includegraphics{images/img_0062.jpg}}
\newlength{\imageheightd}
\settoheight{\imageheightd}{\includegraphics{images/img_0062.jpg}}
	 \includegraphics[trim=0.15\imagewidthd{} 0.55\imageheightd{} 0.8\imagewidthd{} 0.35\imageheightd{}, clip, width=0.49\textwidth]{images/img_0062.jpg}
	 \includegraphics[trim=0.8\imagewidthd{} 0.6\imageheightd{} 0.1\imagewidthd{} 0.2\imageheightd{}, clip, width=0.49\textwidth]{images/img_0062.jpg}
\mbox{Count: 2638}
\end{subfigure}
\begin{subfigure}[b]{0.33\textwidth}
\centering
	\includegraphics[width=\textwidth]{images/bayesian_img_0062_mae-401.72_gt-2638_pred-2236.28_psnr-30.654_ssim-0.079.png}
	\newlength{\imagewidthe}
     \settowidth{\imagewidthe}{\includegraphics{images/bayesian_img_0062_mae-401.72_gt-2638_pred-2236.28_psnr-30.654_ssim-0.079.png}}
     \newlength{\imageheighte}
\settoheight{\imageheighte}{\includegraphics{images/bayesian_img_0062_mae-401.72_gt-2638_pred-2236.28_psnr-30.654_ssim-0.079.png}}
	 \includegraphics[trim=0.15\imagewidthe{} 0.55\imageheighte{} 0.8\imagewidthe{} 0.35\imageheighte{}, clip, width=0.49\textwidth]{images/bayesian_img_0062_mae-401.72_gt-2638_pred-2236.28_psnr-30.654_ssim-0.079.png}
	 \includegraphics[trim=0.8\imagewidthe{} 0.6\imageheighte{} 0.1\imagewidthe{} 0.2\imageheighte{}, clip, width=0.49\textwidth]{images/bayesian_img_0062_mae-401.72_gt-2638_pred-2236.28_psnr-30.654_ssim-0.079.png}
	
	\mbox{Count: 2236.3, PSNR: 30}
	\end{subfigure}
\begin{subfigure}[b]{0.33\textwidth}
\centering
     \includegraphics[width=\textwidth]{images/our_img_0062_mae-18.68_gt-2638_pred-2656.68_psnr-37.114_ssim-0.255.png}
     \newlength{\imagewidthf}
     \settowidth{\imagewidthf}{\includegraphics{images/our_img_0062_mae-18.68_gt-2638_pred-2656.68_psnr-37.114_ssim-0.255.png}}
     \newlength{\imageheightf}
\settoheight{\imageheightf}{\includegraphics{images/our_img_0062_mae-18.68_gt-2638_pred-2656.68_psnr-37.114_ssim-0.255.png}}
	 \includegraphics[trim=0.15\imagewidthf{} 0.55\imageheightf{} 0.8\imagewidthf{} 0.35\imageheightf{}, clip, width=0.49\textwidth]{images/our_img_0062_mae-18.68_gt-2638_pred-2656.68_psnr-37.114_ssim-0.255.png}
	 \includegraphics[trim=0.8\imagewidthf{} 0.6\imageheightf{} 0.1\imagewidthf{} 0.2\imageheightf{}, clip, width=0.49\textwidth]{images/our_img_0062_mae-18.68_gt-2638_pred-2656.68_psnr-37.114_ssim-0.255.png}
	
     \mbox{Count: 2656.7, PSNR: 37}
     \end{subfigure}

\caption{{\bf Density maps} produced by the proposed method. 
\label{fig:visualization}}
\end{figure*}

\section{Conclusion}
In this paper, we show that using Gaussian kernel to smooth the ground truth dot annotation can hurt the generalization bound of a model when testing on the real ground truth data. Instead, we consider crowd counting as a distribution matching problem and propose DisCount to address this problem. Unlike previous works, DisCount does not need any Gaussian kernel to smooth the annotated dots. We present the generalization error bounds of DisCount is tighter than that of the Gaussian smoothed methods. Extensive experiments on four crowd counting benchmarks demonstrate that DisCount significantly outperforms previous state-of-the-art methods.


\section*{Broader Impact}

Our work is able to more accurately estimate the crowd count in images or videos, such that it can guide crowd control and improve public safety. Also, potentially it could be used to protect public health by monitoring social distance which is becoming increasingly important during epidemic. The social public is a biggest beneficiary group from our work. This method does not leverages bias in the data. 



{\small
\bibliographystyle{plainnat}
\bibliography{shortstrings,pubs,egbib}
}